# The LALA Benchmark: Evaluating Literary and Linguistic Analysis in LLMs

Michael Foster[1] Barbara McGillivray[2] Gard B. Jenset[3]
Jakub Mácha[4] Jason Harding[5] Marcus Perlman[1] Devyani Sharma[6]
Patrick Skeels[7] Jana Willer-Gold[6] Jeannette Littlemore[8]

[1] LALA Labs [2] King's College London [3] Ghent University
[4] Masaryk University [5] Lingnan University [6] University of Oxford
[7] University of Kentucky [8] University of Birmingham

Correspondence: michael@lalalabs.ai

## Abstract

We introduce the Literature and Linguistics Analysis Benchmark (LALA), comprising 236 open-ended questions written by experts in linguistics and literary studies, together with expert-provided ideal answers. By comparing models within families and at similar training scales using human and LLM judges, LALA identifies differences in linguistic competence associated with post-training, offering a framework for evaluating future models and guiding targeted improvements. We administered questions to 17 language models and evaluated their responses using both human expert assessment and two LLM evaluators, Sol and Qwen. Base models displayed little of the linguistic competence assessed by the benchmark, whereas their post-trained versions performed markedly better, suggesting that post-training plays a central role in making linguistic knowledge accessible and usable. Parameter count, by contrast, showed no statistically significant association with partial-credit performance, and competencies remained unevenly distributed across linguistic domains. We interpret this pattern as evidence that post-training adds linguistic competencies piecemeal, without producing a general linguistic intelligence capable of extending them beyond the learned policies that support them. Models performed best on logical relations between statements and selected discourse-level tasks, while even the strongest post-trained models continued to struggle with ambiguity, literary style, reference, and misleading assumptions. Human and Sol evaluations produced nearly identical model rankings in both the original and ReliableEval runs, supporting the replicability of the comparative results.

## 1 Introduction

The Literature and Linguistics Analysis Benchmark (LALA) measures how well large language models can analyze language by assessing how closely their responses match the competency of expert linguists and literary analysts.

The two main contributions of this paper are:

A. A description of the methodology we followed to design, build, and evaluate the benchmark.
B. A quantitative evaluation of seventeen Large Language Models (LLMs) measured against the benchmark.

The benchmark consists of 236 questions, written by linguists and literary scholars, covering linguistic analysis, textual interpretation, and logical inference and entailment in natural language. Each question is accompanied by a gold standard ideal response supplied by its author and used to score the models' answers. For the evaluation, we prompted seventeen LLMs with the 236 questions from the benchmark, and humans assessed their responses. A separate run based on the ReliableEval method of rewording questions was used to examine the reproducibility of the findings and provide a check on the potential subjectivity of human assessment (Lior et al., 2025). Two language models, GPT-5.6 Sol and Qwen3 14B, assessed the responses from this run using the same rubric as the human assessors. We compared available open-weight models of various parameter counts. These served as proxies for differences in training, while base and post-trained versions within the DeepSeek and Mistral families allowed closer comparisons between related models. The panel also included proprietary state-of-the-art models from Anthropic, OpenAI, and Google.

The aims of the LALA benchmark are as follows:

1. Assess to what extent the models can infer expected linguistic competencies as humans naturally do and apply them to the analysis of various linguistic utterances and textual artefacts, leveraging recognized subtle distinctions across semantic, syntactic, and pragmatic aspects of natural language use.
2. To investigate how linguistic competency differs between base models and models that have undergone different forms of post-training.

With reference to our aims, our central findings are:

1. Models display varying degrees of linguistic and literary competency in particular tasks, but these gains do not reliably generalize to other kinds of linguistic analysis. They may handle logical inference while still missing ambiguity, changes in reference, or literary meaning. We interpret this as evidence that the models are given specific response policies without acquiring a general ability to interpret and analyze language or a generalizable linguistic intelligence.
2. Post-trained models answered questions in the LALA benchmark substantially better than the base models. Both base models tested were unable to produce almost any of the linguistic competency under assessment. The stronger models performed relatively well on logical inference and conversational structure. Their weaknesses included literary framing, stylistics, reference, and misleading assumptions. Humans and Sol ranked the models almost identically and identified several of the same strengths and weaknesses. Overall, post-training seems to add linguistic competencies piecemeal, and the models show little ability to infer linguistic ability beyond added competencies. Together, these findings help identify which aspects of linguistic competency are still not being injected into the models via post-training.

## 2 Related work

Many linguistic benchmarks test specific grammatical distinctions. For example, BLiMP uses minimal sentence pairs to test morphology, syntax, and semantics (Warstadt et al., 2020), while SyntaxGym provides controlled test suites grounded in psycholinguistic experimental design (Gauthier et al., 2020). MultiBLiMP 1.0 extends minimal-pair evaluation to 101 languages, testing agreement through sentence probabilities (Jumelet et al., 2025). These methods measure sensitivity to linguistic structure without requiring an explanation. In contrast, LALA asks models to articulate the distinctions on which their judgments depend, thus assessing their explicit understanding of language.

Other benchmarks directly test reasoning about language. IOLBENCH asks models to infer linguistic rules from Olympiad problems and includes explanatory answers assessed against expert solutions (Goyal and Dan, 2025). Arčon et al. (2026) convert World Atlas of Language Structures features into multiple-choice questions about the grammatical properties of 2,660 languages. The former tests rule induction; the latter tests explicit knowledge about languages. LALA attempts to target the models' ability to respond like an expert when given opaque or confusing prompts that rely on a developed ability to comprehend and respond to complex sentences and passages that demand a high level understanding of language in use.

ALBA is a close comparison in both authorship and answer format. Its 800 expert-written questions cover eight dimensions of European Portuguese, including discourse, wordplay, syntax, and phonology, and its automated judge is calibrated against expert-rated responses (Vieira et al., 2026). ALBA combines linguistic analysis with constrained generation and language-variety tasks. LALA concentrates on the abilities of models to analyze linguistic distinctions, contextual meaning, and literary form.

Literary benchmarks address several abilities related to each other. KRISTEVA tests close reading through questions connecting style, context, and interpretive reasoning (Sui et al., 2025). LiteraryQA evaluates generated answers to questions over complete books (Bonomo et al., 2025). PersLitEval evaluates the analysis of Persian literature using an examination of multiple-choice questions covering literary devices, conceptual understanding, and formal linguistic analysis (Niazi et al., 2026). EduArt assesses LLMs' ability to analyze questions in the domain of art history (Spinaci et al., 2026). LALA joins this work on literary benchmarks through questions that ask models to develop an interpretation and address the points identified in an expert's ideal answer.

Recent pragmatic benchmarks also evaluate explanations. AltPrag asks models to infer speaker intentions and explain when and why one utterance would be chosen over a plausible alternative (Yu et al., 2025). VIVID evaluates explanations of Vietnamese idioms and proverbs as well as classification tasks (Do et al., 2026). Like these tasks, LALA requires an explanation of meaning, while also covering formal linguistic and literary analysis.

Other tests examine when a model should resist an inference. PaCE distinguishes contexts that license a pragmatic inference from those in which it becomes an unwarranted addition (Li et al., 2026). FLEX tests whether responses reject false presuppositions in German political questions (Sieker et al., 2025). Hong et al. (2026) test situated social meaning through human-validated multiple-choice items based on Chinese online exchanges, with distractors representing plausible misreadings. These studies bear on LALA's questions about misleading premises, social meaning, and ambiguity, where a successful answer may need to challenge an assumption or preserve more than one reading.

Instruction tuning can improve the usefulness of model responses (Ouyang et al., 2022), but linguistic effects vary with the task. AltPrag reports pragmatic sensitivity in base models and further gains after post-training (Yu et al., 2025), whereas MultiBLiMP finds that instruction tuning can reduce performance on grammatical minimal pairs (Jumelet et al., 2025). These findings make task and elicitation differences central to interpreting LALA's base and post-trained comparisons. Hu and Levy (2023) found that models could recognize grammatical distinctions more reliably than they could answer explicit questions about them. We therefore document the evaluation conditions, following Hu and Levy's (2023) distinction between prompted judgments and probability measurements.

Language models have shown promising agreement with human judges in some settings, but full correspondence in LLM judgment with human judgment remains elusive (Zheng et al., 2023). To address this, we compared human and automated judgments through model rankings and scores across linguistic categories. These comparisons address two related questions: whether the evaluation procedures produce a similar ordering of models, and whether they identify similar patterns of strengths and weaknesses. Agreement in rankings would support the stability of our conclusions about relative model performance. Similar category-level patterns would provide further evidence that those conclusions extend to that level of analysis, while accounting for smaller differences in judgment at the level of individual question.

Because humans and the automated judges assessed separate response sets, this comparison measures consistency across evaluation procedures. Any observed differences may reflect variation in the answers themselves as well as differences between judges. We therefore do not directly measure how often a human and an automated judge would assign the same score to the same answer, or whether they would recognize the same errors and interpretive merits. Establishing that form of agreement would require both to assess a shared set of responses under the same rubric.

## 3 Benchmark construction

### 3.1 Questions and reference answers

The benchmark consisted of 236 questions, which were contributed by seven linguists, two philosophers and two literary scholars. The questions were written in English. A question about a logical inference might require showing how a conclusion follows from the premises and identifying a mistake in the question. An ambiguous sentence might require explaining its possible meanings and why the context does not resolve the ambiguity. A literary question might require recognizing an allusion or explaining how a change in voice affects the meaning of a passage.

For each question, its author supplied an ideal answer and keywords describing the linguistic concepts or interpretive skills being tested. The ideal answers served as the gold standard in the evaluation (section 4).

**Box 1. Two sample prompts from LALA**

**Linguistic prompt (Q-000207)**
Analyse this sentence linguistically: precutatording aferts can be retrous.

**Literary prompt (Q-000103)**
Demonstrate how the meaning of "honest" in *Othello* is resolved into unambiguous praise of the tragic hero

### 3.2 Scope of competencies assessed

LALA places particular emphasis on how context affects meaning. The questions test pragmatic inference, speech acts, and social meaning, as well as the influence of genre, rhetorical form, and changes of voice on the interpretation of a text. They also test whether models recognize ambiguity and can explain what possible readings are available to a reader of the prompt.

We organize the questions under three broad headings: Linguistics, Logic, and Textuality. Linguistics covers formal and lexical semantics, pragmatics, discourse, morphosyntax, phonology, and sociolinguistics. Logic covers the interpretation of logical relationships in ordinary language, including conditionals, quantifiers, negation, and statements about knowledge or possibility. Textuality covers literary analysis, philology, and the ability to interpret a prompt and assess what it asks or assumes about the textual composition of the prompt or text under inquiry.

The questions are tagged with 298 keywords, grouped into 27 categories (see Appendix C):

**Linguistics:** Cognitive Pragmatics, Discourse Analysis, Dynamic Semantics, Formal Semantics, General Linguistics, Lexical Semantics, Linguistic Pragmatics, Metasemantics, Morphosyntax, Phonology, Social Pragmatics, and Sociolinguistics.

**Logic:** Argumentation, Conditional Logic, Epistemic Logic, Epistemology, General Logic, Mathematical Intuition, Modal Logic, Predicate Logic, Probabilistic Logic, Proof Theory, Propositional Logic, and Syllogistic Logic.

**Textuality:** AI Evaluation, Literary Analysis, and Philology.

The categories overlap and are not equally represented in the benchmark. For example, a question about reference may also involve ambiguity or an unsupported presupposition. Category scores reflect performance on the questions as a whole, including any other skills needed to answer them.

### 3.3 What the questions require

The benchmark generally requires models to answer questions covering linguistic analysis, textual interpretation, and logical inference and entailment in natural language. Some questions contain purposeful distortions, including typos and deliberately confusing phrasing. These were included to test how easily such wording confuses models and whether they can still understand the question. When the wording allows more than one interpretation, the complete answer must acknowledge the ambiguity.

The following examples illustrate how ambiguity, stylistic alteration, and misleading phrasing enter the questions. In one question, a speaker says they have a great deal of backstory to explain. The reply, "Drop the lore man," could mean either "tell me the backstory" or "leave out the backstory" (Q-000013). The ideal answer explains both meanings, and full credit requires both readings.

Another question presents Hamlet's soliloquy rewritten in the style and format of an AI chatbot (Q-000215). The model must draw on its knowledge of the original passage and explain how the chatbot style changes it. The task asks for an analysis of the stylistic transformation.

A third question presents a logical argument in which a statement about the past is followed by a conclusion about the future (Q-000184). The model must notice the change in tense and explain why the argument does not follow the logical pattern it appears to use.

**Box 2. Questions and ideal answers across the three main categories**

**Linguistics — Question (Q-000013)**
A: I have so much to tell you, it's going to explain me and I need to give you a ton of backstory.
B: Drop the lore man.
What is B implicitly saying to A?

**Ideal answer**
B is possibly saying two things. Either B means "please tell me what you have to tell me, including all of the backstory that you see as necessary and interesting" or B means "do not be overly verbose and provide me with too much information and contextual information if you want to tell me something," with possibly the valence in B's request akin to a desire that A not overly self-mythologize when providing this information.
In other words, B's statement is underdetermined in this snippet and as a result could have two different meanings and intentions behind it that are functionally opposed to one another.

**Textuality — Question (Q-000099)**
"Excellent wretch! Perdition catch my soul, But I do love thee, and when I love thee not Chaos is come again." Explain how this speech demonstrates Othello's jealous suspicions regarding Desdemona.

**Ideal answer**
This speech is made before Othello has any suspicions about Desdemona. Dr Johnson points out that the meaning of "wretch" here in Elizabethan English is a term of the "softest and fondest tenderness". Othello is therefore expressing extreme tender feelings towards Desdemona. His total and unambiguous endearment is a measure of his absolute love for her at this point in the play.

**Logic — Question (Q-000178)**
Possibility can be characterized in terms of necessity and negation. Here are some examples:
- It's possible that Earth is flat. Therefore, it's not the case that it's necessary that Earth isn't flat.
- It's possible that they didn't make it. Therefore, It's not necessary that that they made it.
Can you give me an example of this inference pattern where the prejacent of the modal is not atomic?

**Ideal answer**
Sure:
- It's possible that Alice passed the exam and Bob failed the exam. Therefore, it's not the case that it's necessary that it's not the case that both Alice passed the exam and Bob failed the exam.

## 4 Evaluation and assessment

### 4.1 Model selection

We evaluated the following models:

- Two Base models with open weights:
    - DeepSeek V3 Base

- Mistral 3 14B Base
- Twelve post-trained models with open weights:
    - DeepSeek V3 and R1
    - Mistral 3 14B Instruct and Reasoning
    - Llama 3.1 8B Instruct and Llama 3.2 3B Instruct
    - Gemma 4 31B and 26B A4B
    - GPT-OSS 120B
    - Qwen3 235B A22B, Qwen3 235B A22B 2507, and Qwen3 235B A22B 2507 Thinking
- Three post-trained proprietary models:
    - Claude Fable 5
    - Gemini 3.1 Pro Preview
    - GPT-5.6 Sol

The open models ranged from roughly 3 billion to 671 billion total parameters. In mixture-of-experts models, only a portion of these parameters is active during each computation. Parameter counts for proprietary models are not publicly available.

To help examine the role of post-training, we compared base models with post-trained versions from the same family. The Ministral models each had 14 billion parameters, and DeepSeek provided a similar comparison at a much larger scale.

### 4.2 The original run and the ReliableEval run

In the original run, each model received one question at a time, without the gold standard answer, scoring rubric, worked examples, or additional explanation. We used the completion or chat interface and instruction and reasoning formats required by each model. System instructions varied across models. Humans assessed the answers from this run.

For the ReliableEval run, models answered several reworded versions of each question, following the approach of Lior et al. (2025). Qwen3 14B generated 2,128 of the 2,131 prompt variants according to ReliableEval's proposed method, while Q-0002050 and Q-000249 were rewritten according to a fixed rule to reverse adjacent capitalized words joined by "and" after the model repeatedly failed to produce rewrites for these two questions. The model was instructed to vary the wording while preserving the question's meaning and expected answer. GPT-5.6 Sol and Qwen3 14B assessed the same answers using the contributor's gold standard answer and the rubric used by humans.

The ReliableEval run covered 222 of the 236 questions, with up to ten versions of each, giving 2,131 prompts per model and 36,227 response records. Each record received scores from both automated judges. The other 14 questions had no acceptable reworded versions and were excluded from comparisons across all three assessments.

Missing outputs and responses containing no assessable text received zero in the main results. For Ministral Reasoning, we also recalculated Sol's score after excluding generation failures, as reported in section 5.2. Incorrect and incomplete answers remain included; questions with no assessable responses were omitted from this recalculation. For models with frequent generation failures in response to a given prompt, we also recalculated scores using only responses that contained assessable text. Incorrect and incomplete answers remained in this secondary analysis. Questions for which a model produced no assessable response were omitted from that calculation.

### 4.3 Assessment rubric

Six assessors selected from the authors of the ideal answers in the benchmark were asked to assess all 236 questions and the responses of all 17 models against the ideal answers, yielding 4,012 scores. In their assessment, the human experts followed a rubric covering three criteria:

a. How fully does the response address the gold standard ideal answer?
b. Is the model's response accurate?
c. Is the model's response relevant to the question?

Each criterion was scored from 0 to 2. 0 indicates little or no successful analysis, 1 indicates substantial analysis that is incomplete or partly incorrect. A score of 2 indicates a satisfactory analysis covering the required points; wordings equivalent to the ideal answer were accepted. For literary questions with several reasonable interpretations, assessors judged whether a response addressed the points the author considered necessary.

For answers with an overall score of 1, assessors also scored the skills represented by the question's keywords. These scores recorded which parts of the expected analysis the answer supplied. Credit required explaining the relevant concept: an answer that mentioned "irony," for example, had to explain the ironic meaning. The keyword scores therefore give a more detailed account of partially successful answers, with each skill judged in the context of the whole question.When the overall score was 0 or 2, every keyword received the same score.

The models' responses were also assessed by "automated judges": Qwen3 14B and GPT-5.6 Sol, referred to below as Qwen and Sol. The Qwen3 14B judge was separate from the Qwen3 235B A22B models tested as respondents. Both judges received the rubric as instructions; neither was fine-tuned for this task. Sol used an extended reasoning setting.

### 4.4 Scoring and comparisons

We calculated an overall score for each model as percentages of the maximum rubric score, which we call normalized rubric attainment. We divided the mean score by 2 and expressed the result as a percentage. A score of 50% corresponds to a mean of 1 on the 0–2 scale and includes partial credit. For the ReliableEval run, we first averaged each model's scores across the versions of a question, then averaged across questions. Every original question thus contributed equally to the model's overall score.

Unless otherwise stated, comparisons between human and model scores used the same 222 questions covered by the ReliableRun (section 4.2). When recalculating scores after excluding missing outputs whether from refusals or generation failures, we report how the questions were selected and how many remained in each analysis (see section 5.2 for exclusion criteria and section 5.7 for human-score comparisons. To calculate category scores, we selected questions with at least one keyword in the category and averaged their overall scores. Each question counted once within a category, and each model carried equal weight. A question's score covered the full answer, including any skills outside the category.

We defined question difficulty by the mean human-assessed score across the 17 models, with lower scores indicating greater difficulty. We ranked all 236 questions by this mean and divided them into four equal groups of 59, labelled hardest, hard, moderate, and easiest; ties were ordered by question ID. These groups describe relative difficulty for the models evaluated under our rubric.

For selected comparisons within model families, we calculated descriptive 95% bootstrap intervals using 5,000 samples drawn with replacement from the questions. Each sample kept the two models' scores for a question together. The intervals show how the estimated score difference varied with the selection of questions. Differences in training and evaluation conditions also need to be considered when interpreting these comparisons.

For Figure 2, we calculated descriptive 95% confidence intervals for each model's mean using 5,000 bootstrap samples of questions drawn separately within the 236-question full set and the 222-question shared set. Bounds are the 2.5th and 97.5th percentiles of the resampled means, converted to percentages of the maximum rubric score.

We compared model means using Pearson's r and model rankings using Spearman's ρ and Kendall's W. Analyses used unrounded mean scores on the 222 shared questions for all 17 models, with a further Human–Sol comparison excluding the two base models ($n = 15$). There were no tied model means. For two rankings without ties, $W = (1 + \rho)/2$: $W = 1$ indicates identical rankings, $W = 0.5$ corresponds to zero rank correlation, and $W = 0$ indicates reversed rankings (Legendre, 2010). We also calculated W across all three judges.

We calculated unadjusted Monte Carlo permutation p-values using 999,999 random model-label shuffles per comparison. For each pair of judges, one score vector was held fixed while the other was permuted; for three-judge W, Sol and Qwen scores were permuted independently while human scores were held fixed. Tests were two-sided for r and ρ, using absolute coefficients, and upper-tailed for W. We used $p = (b + 1)/(999{,}999 + 1)$, where b counts permutations at least as extreme as the observed statistic. These p-values do not require normally distributed scores. They test independent model-label pairing, conditional on the evaluated models and their observed scores.

We compared overall model scores and rankings, as well as scores for individual questions. In Human–Sol and Human-Qwen comparisons, the human score for each original answer was paired with the average score for that model’s answers to the ReliableEval versions of the same question. Sol and Qwen scored identical answers from the ReliableEval run, allowing us to compare their judgments on each response directly.

## 5 Results

### 5.1 Overall performance

The three proprietary models led both the human and Sol rankings. Claude Fable 5 ranked first, with an overall accuracy of 72.3% under humans and 62.0% under Sol. Gemini 3.1 Pro Preview scored 64.4% and 52.9%, and GPT-5.6 Sol scored 59.9% and 56.0%. Humans placed Gemini second and Sol third; Sol's assessment reversed their order. Gemma 4 31B ranked fourth in both assessments, at 51.1% and 41.6%, followed by DeepSeek R1 at 49.1% and 38.1%.

Sol's higher ranking of itself raises the possibility of self-preference bias. Wataoka et al. (2024) found that GPT-4 favored its own responses in comparisons with human judgments. Spiliopoulou et al. (2025) found that GPT-4o and Claude 3.5 Sonnet favored their own outputs and those of other models in the same family, even after accounting for response quality using human ratings. In LALA, humans assessed answers from the original run and Sol assessed answers from the ReliableEval run. Differences between those answers could also explain the change in ranking, so the extent of any self-preference remains uncertain.

Gemma 4 31B outscored the much larger DeepSeek and Qwen models under both human and Sol assessment. The 671-billion-parameter DeepSeek base model was near the bottom. Performance therefore did not increase consistently with parameter count. Architecture and training data also differed between models and may have contributed to their scores.

Even the leading models received partial credit on many questions. Their answers often omitted part of the required analysis or contained errors. The category results show where these difficulties were most common. Figure 1 shows the full human score distribution for each model.

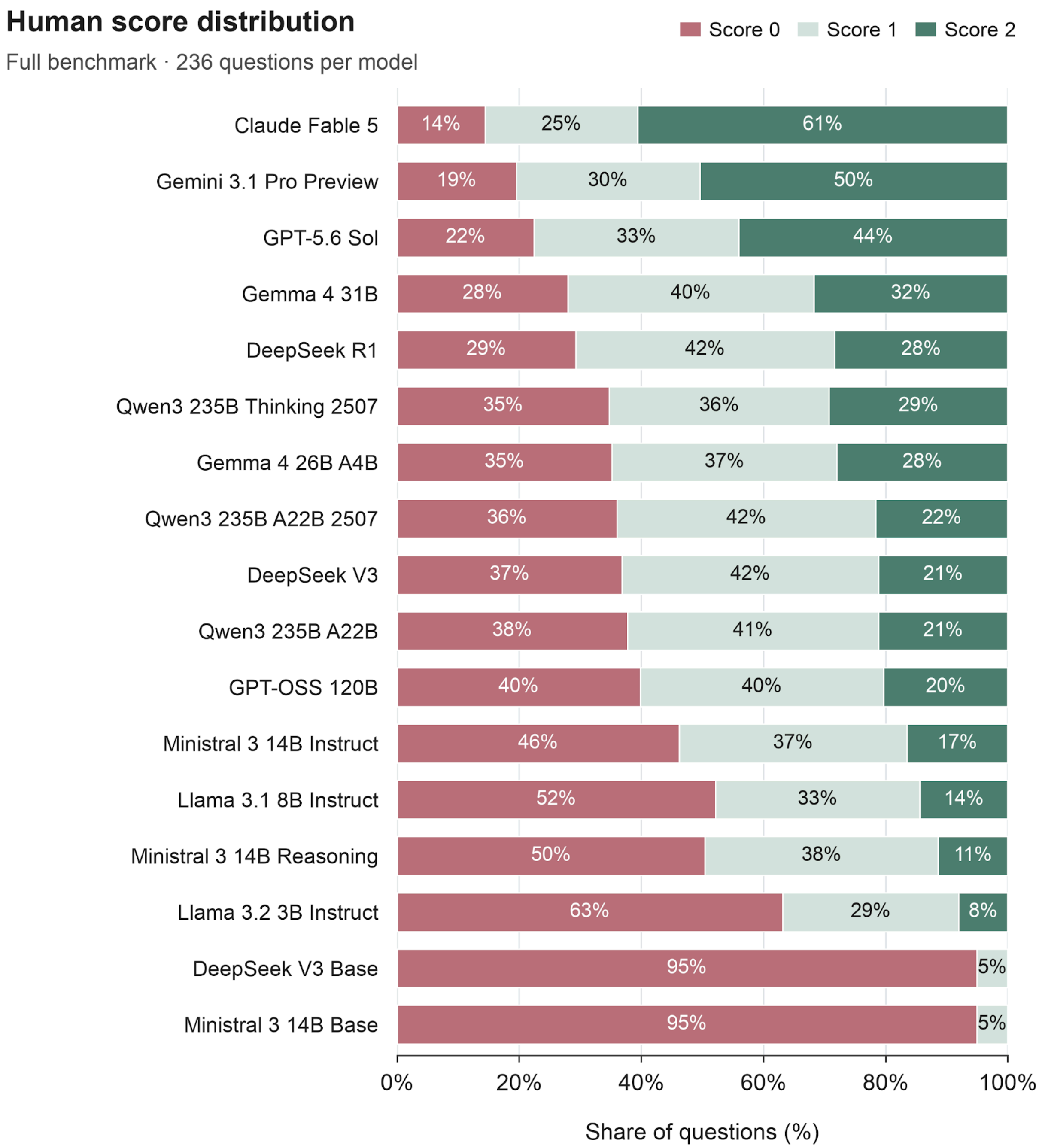

**Figure 1.** Human score distributions for all 17 models across the full 236-question benchmark. Each bar totals 100%; colored segments show the percentages assigned rubric scores of 0, 1, or 2. Labels are rounded. Missing answers received zero. Figure 2 shows mean scores and 95% confidence intervals for both the full set and the 222 shared questions.

Figure 2 compares human means and confidence intervals for the full and shared question sets. Figure 3 shows Sol's results on the shared set.

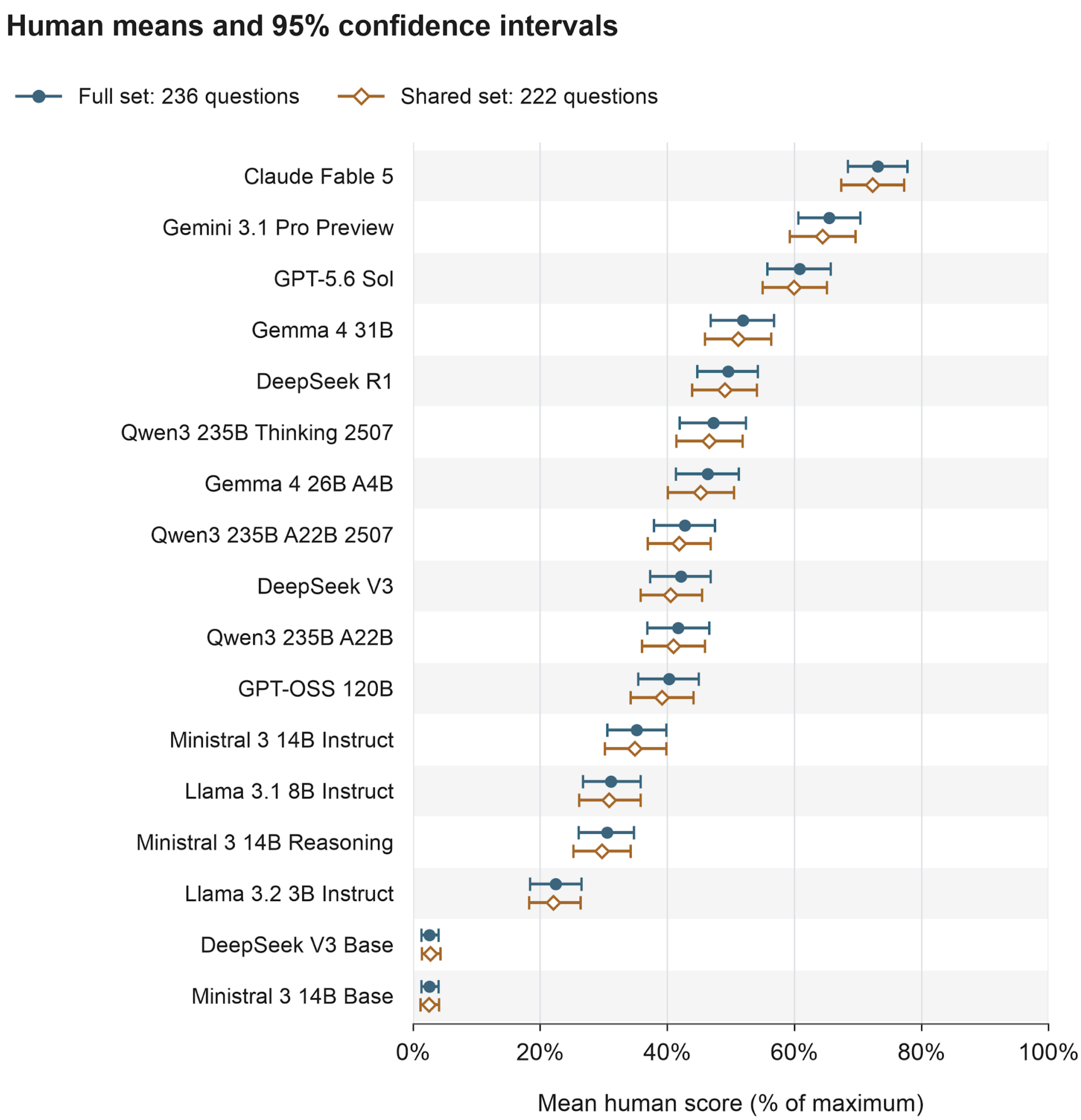

**Figure 2.** Mean human scores for all 17 models on the full 236-question benchmark (filled circles) and the 222 questions shared with ReliableEval (open diamonds). Scores are percentages of the maximum rubric score (mean score on the 0–2 scale × 50). Capped horizontal bars show descriptive 95% percentile bootstrap confidence intervals from 5,000 resamples of questions within each set. Missing answers receive zero; both series use the original human-assessed answers.

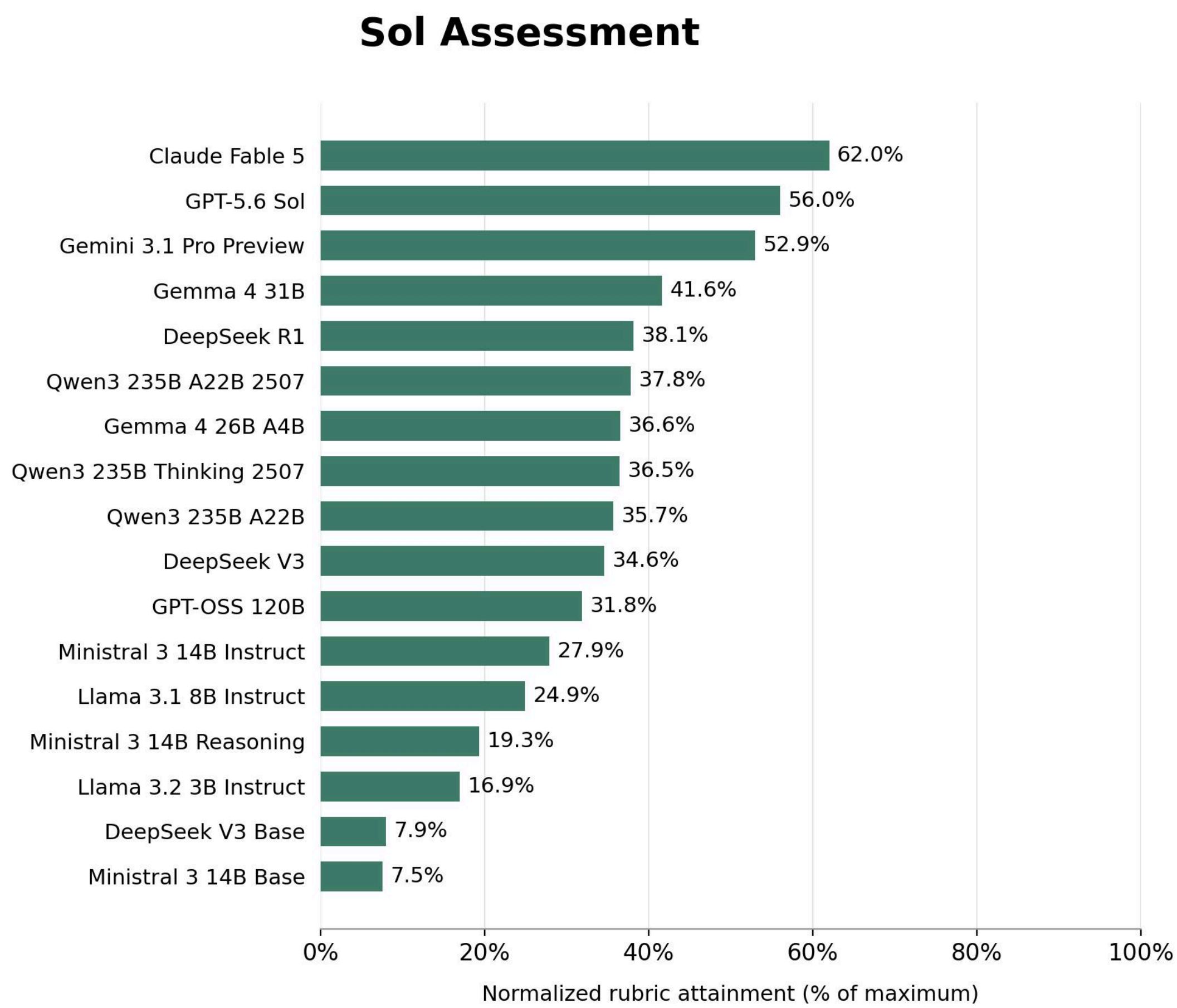

**Figure 3.** Headline GPT-5.6 Sol assessment results for all 17 models on the same 222 questions. Scores were averaged across ReliableEval versions within each question, then equally across questions, and expressed as percentages of the maximum rubric score. Failed responses receive zero. Humans and Sol assessed different answers.

### 5.2 Base models and post-trained models

The two base models ranked last under human, Sol, and Qwen assessments. On the common question set, Mistral 3 14B Base scored 2.5% under humans and 7.5% under Sol. Mistral Instruct scored 34.9% and 27.9%, respectively. The gaps were 32.4 percentage points under human assessment and 20.4 points under Sol, with bootstrap intervals of 27.7–37.2 and 17.8–23.0 points. The two models had the same parameter count, so model size cannot explain this difference.

DeepSeek showed a similar improvement. V3 Base scored 2.7% under humans and 7.9% under Sol, compared with 40.5% and 34.6% for V3. The gap under human assessment was 37.8 percentage points, with a bootstrap interval of 33.1–43.0; under Sol, it was 26.6 points, with an interval of 23.5–29.5. DeepSeek R1 scored higher still, at 49.1% and 38.1%. In both families, post-trained models demonstrated substantially greater linguistic competency than their base versions.

**Table 1.** Base and post-trained model scores on the 222 shared questions.

| Model | Human (%) | Sol (%) |
|---|---|---|
| Ministral 3 14B Base | 2.5 | 7.5 |
| Ministral 3 14B Instruct | 34.9 | 27.9 |
| Ministral 3 14B Reasoning | 29.7 | 19.3 |
| DeepSeek V3 Base | 2.7 | 7.9 |
| DeepSeek V3 | 40.5 | 34.6 |
| DeepSeek R1 | 49.1 | 38.1 |

Scores are percentages of the maximum rubric score. Humans assessed the original response set; Sol assessed the separate ReliableEval response set. Missing outputs and generation failures contribute zero.

Both base models returned answers for every question in the ReliableEval run. They received low scores because their answers rarely supplied the linguistic analysis required by the questions. Prompting and output limits also differed between base and post-trained models, as discussed in Section 7.

Mistral Reasoning scored 29.7% under humans and 19.3% under Sol. Failures to produce an answer affected 35.5% of its records in the ReliableEval run. Excluding these failures left 1,374 responses across 199 questions; every version failed on the other 23 questions. On the 199 questions with usable answers, Sol awarded 29.8% to Reasoning and 28.3% to Instruct. This separate calculation shows how much missing output lowered Reasoning's score in the main comparison. Figure 4 compares the base and post-trained models using human scores across all 236 questions.

**Table 2.** Effect of excluding generation failures from Ministral Reasoning scores.

| Model | Sol: all 222 questions (%) | Sol: 199-question subset (%) |
|---|---|---|
| Ministral 3 14B Reasoning | 19.3 | 29.8 |
| Ministral 3 14B Instruct | 27.9 | 28.3 |

The 199-question analysis omits Reasoning’s 757 failed responses and 23 questions with no usable response, leaving 1,374 responses. Instruct is recalculated on the same 199 questions using all its responses. Questions are weighted equally; scores are percentages of the maximum.

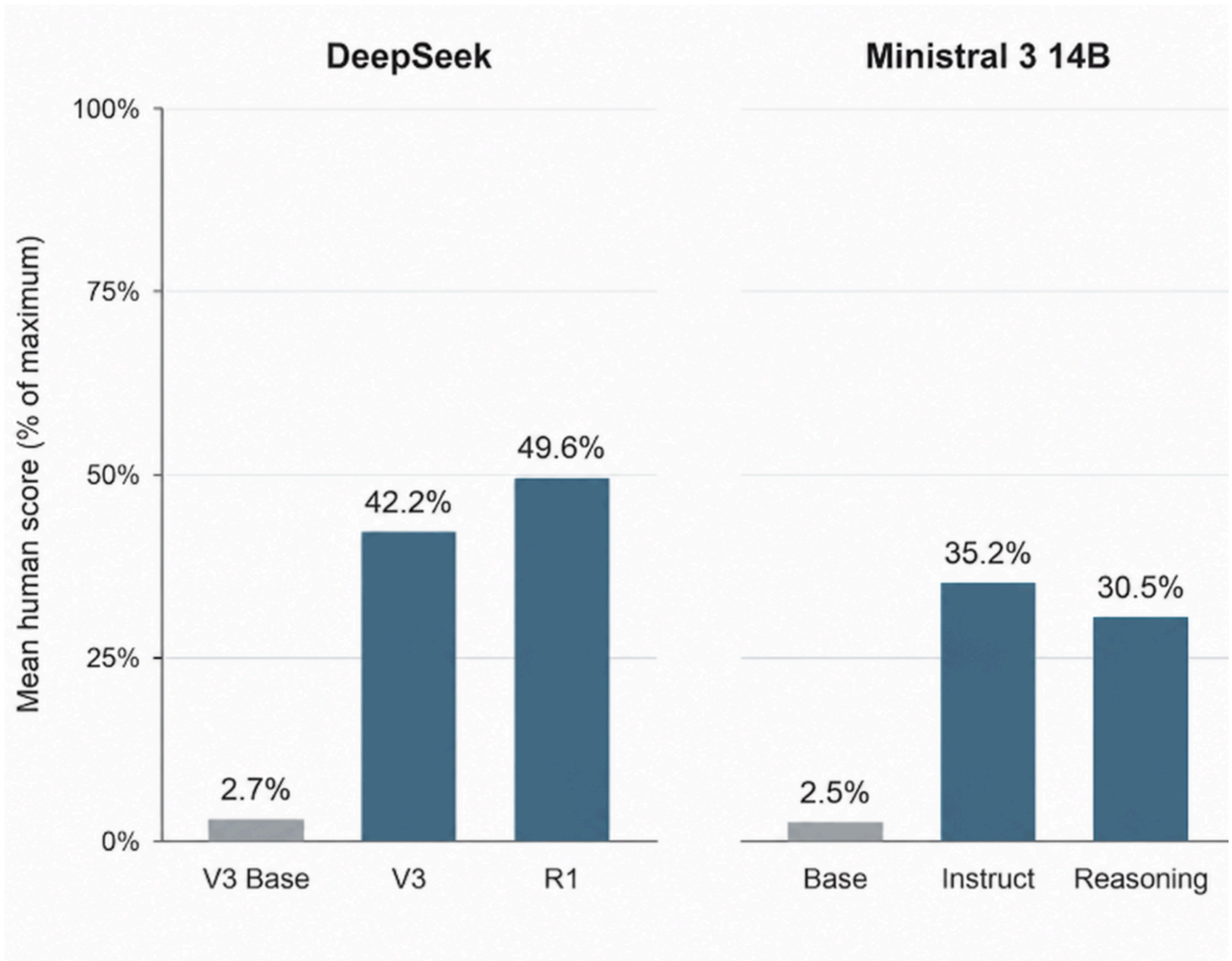

**Figure 4.** Human scores for the DeepSeek and Ministral families across all 236 questions in the original run. Bars show mean scores as percentages of the maximum rubric score. Both base models scored far below their post-trained versions. The comparisons in the text use the 222 questions shared with the ReliableEval run.

### 5.3 Linguistic strengths and weaknesses

Humans and Sol gave the same ordering to the three broad categories of questions. Mean scores were highest for Logic at 45.3% under humans and 38.7% under Sol, followed by Linguistics at 39.2% and 33.6%, and Textuality at 36.9% and 28.8%. Table 3 shows human and Sol scores across the three broad categories and the 27 subcategories.

Among the more detailed categories, Propositional Logic scored 48.1% under humans and 46.1% under Sol across 39 questions. Argumentation scored 45.1% and 43.7% across 37 questions. Models were comparatively successful at identifying and explaining logical relationships between statements.

Discourse Analysis was another relative strength, with scores of 41.9% under humans and 41.8% under Sol across 23 questions. Models were less successful at interpreting particular expressions and literary passages.

Literary Analysis scored 35.1% under humans and 26.0% under Sol across 40 questions; Lexical Semantics scored 31.3% and 28.8% across 37 questions. The 12 Metasemantics questions, covering use, mention, quotation, and self-reference, scored 37.7% and 23.5%. AI Evaluation, which includes misleading premises and deliberately altered wording, scored 25.4% and 21.0% across 20 questions.

These questions often exposed a failure to check a familiar interpretation against the words of the passage. Figure 5 shows the distribution of question difficulty across all 27 specific categories in the full benchmark.

**Table 3.** Human and Sol scores across all broad categories and subcategories.

| Category | Questions | Human (%) | Sol (%) |
|---|---|---|---|
| LINGUISTICS | 177 | 39.2 | 33.6 |
| Linguistic Pragmatics | 46 | 37.1 | 30.6 |
| Social Pragmatics | 41 | 42.3 | 34.2 |
| Morphosyntax | 38 | 38.3 | 35.1 |
| Formal Semantics | 37 | 38.9 | 34.8 |
| Lexical Semantics | 37 | 31.3 | 28.8 |
| Sociolinguistics | 27 | 37.6 | 35.5 |
| Phonology | 24 | 44.2 | 38.3 |
| Discourse Analysis | 23 | 41.9 | 41.8 |
| Metasemantics | 12 | 37.7 | 23.5 |
| Dynamic Semantics | 8 | 42.3 | 37.1 |
| General Linguistics | 6 | 46.1 | 38.3 |
| Cognitive Pragmatics | 2 | 35.3 | 8.1 |
| TEXTUALITY | 96 | 36.9 | 28.8 |
| Philology | 47 | 40.3 | 31.2 |
| Literary Analysis | 40 | 35.1 | 26.0 |
| AI Evaluation | 20 | 25.4 | 21.0 |
| LOGIC | 77 | 45.3 | 38.7 |
| Propositional Logic | 39 | 48.1 | 46.1 |
| Argumentation | 37 | 45.1 | 43.7 |
| Modal Logic | 18 | 51.6 | 41.9 |
| Conditional Logic | 11 | 49.5 | 38.8 |
| Epistemic Logic | 7 | 45.0 | 35.5 |
| General Logic | 7 | 43.3 | 32.6 |
| Syllogistic Logic | 7 | 38.2 | 25.0 |
| Epistemology | 6 | 37.7 | 18.7 |
| Predicate Logic | 3 | 52.0 | 42.9 |
| Probabilistic Logic | 3 | 69.6 | 46.1 |
| Proof Theory | 3 | 47.1 | 44.7 |
| Mathematical Intuition | 2 | 27.9 | 20.7 |

Scores are percentages of the maximum, averaged across 17 models on the 222 shared questions. The first three rows are broad categories, followed by all 27 subcategories in alphabetical order. Categories overlap.

The chatbot-style Hamlet question illustrates the LLMs' difficulty with issues of literary analysis and textual style. All 17 answers from the original run received zero from humans. Of 170 responses from the ReliableEval run, Sol gave zero to 168 and partial credit to two; none received full credit. Models failed to explain adequately how rewriting Hamlet's soliloquy as a chatbot response changed its meaning and created a parody.

Models had more success with the question about a shift from past to future tense. Human scores averaged 50.0% and Sol scores 52.6%. Most answers received partial credit: 13 of 17 original responses and 145 of 170 ReliableEval responses scored 1. Models often recognized the logical pattern and left the change in tense unexplained.

Sol's separate scores for each rubric criterion help explain the low overall scores (see Table 4). In Literary Analysis, relevance averaged 83.7%, coverage of the ideal answer 27.6%, and accuracy 36.9%. The corresponding values for AI Evaluation were 85.1%, 22.4%, and 44.8%. The low coverage scores reflect frequent omissions from answers that were otherwise relevant to the question.

**Table 4.** Sol scores by rubric criterion for selected categories.

| Category | Coverage (%) | Accuracy (%) | Relevance (%) |
|---|---:|---:|---:|
| Literary Analysis | 27.6 | 36.9 | 83.7 |
| AI Evaluation | 22.4 | 44.8 | 85.1 |

Mean percentages of the maximum criterion score across 17 models, using the shared-question subset: Literary Analysis (40 questions) and AI Evaluation (20 questions). Coverage measures how fully a response addresses the ideal answer.

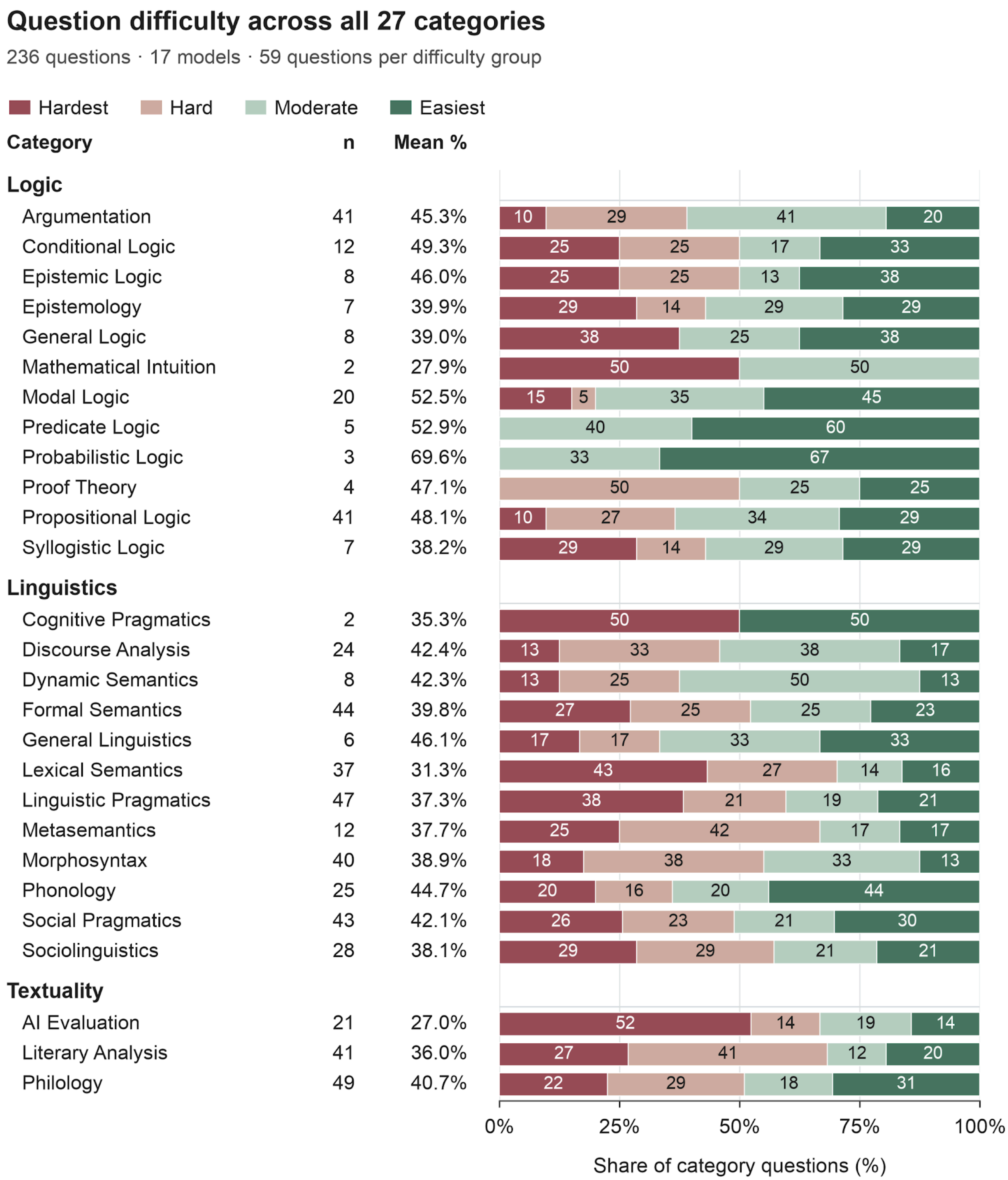

**Figure 5.** Question difficulty across all 27 specific categories in the full 236-question benchmark. Questions were ranked by their mean human score across 17 models and divided into four groups of 59, from hardest to easiest; ties were ordered by question ID. Each bar totals 100%; segment labels give rounded percentages of the category's questions in each group. n gives the number of questions; Mean % gives the mean score as a percentage of the maximum rubric score. Categories overlap.

### 5.4 Agreement between human and GPT-5.6 Sol assessment

Humans and Sol ranked the models almost identically. Across the 17 models, Pearson's r = 0.979, Spearman's ρ = 0.988, and Kendall's W = 0.994 (all $p < 0.001$; Table 5). Thirteen models held the same rank, and none moved more than two places. Humans and Sol agreed on which model scored higher in 132 of the 136 possible model pairings (97.1%).

After removing the two base models, agreement remained close across the other 15 models: Pearson's r = 0.982, Spearman's ρ = 0.982, and Kendall's W = 0.991 (all $p < 0.001$; Table 5). Humans and Sol also placed the broad categories of Logic, Linguistics, and Textuality in the same order. Across all 27 categories, Spearman's ρ = 0.878. The rankings of narrower categories varied more, particularly those represented by few questions.

**Table 5.** Agreement between model scores and rankings across judges.

| **Comparison** | **Pearson r** | **Spearman ρ** | **Kendall W** |
|---|---|---|---|
| Human–Sol (17 models) | 0.979 ($p < 0.001$) | 0.988 ($p < 0.001$) | 0.994 ($p < 0.001$) |
| Human–Sol (15 non-base models) | 0.982 ($p < 0.001$) | 0.982 ($p < 0.001$) | 0.991 ($p < 0.001$) |
| Human–Qwen (17 models) | 0.928 ($p < 0.001$) | 0.819 ($p < 0.001$) | 0.909 ($p < 0.001$) |
| Sol–Qwen (17 models) | 0.883 ($p < 0.001$) | 0.797 ($p < 0.001$) | 0.898 ($p < 0.001$) |
| Human–Sol–Qwen (17 models) | — | — | 0.912 ($p < 0.001$) |

Coefficients are followed by unadjusted permutation p-values in parentheses (999,999 shuffles; see section 4.4). Correlations use model means on the 222 shared questions. Humans assessed original answers; Sol and Qwen assessed the same ReliableEval answers. The final row reports concordance across all three rankings; dashes indicate that a pairwise coefficient does not apply.

Sol generally awarded lower scores than human judges. The human mean across the matched set was 39.7%, compared with 33.4% for Sol, a difference of 6.2%. When we compared each model's score on each question across the two runs, the descriptive Pearson correlation was r = 0.619 and the mean absolute difference was 0.423 on the 0–2 scale. Scores differed by one point or more in 20.7% of these comparisons. Because the two runs produced separate answers, some differences may have come from the answers themselves. Having humans and Sol assess the same answers would show how much disagreement comes from scoring.

### 5.5 Qwen3 14B as a second automated judge

Qwen ranked Fable first and the base models last, although its remaining rankings were less consistent with human assessment. Human–Qwen agreement was Spearman's ρ = 0.819 and Kendall's W = 0.909. Sol–Qwen agreement was Spearman's ρ = 0.797 and Kendall's W = 0.898 (all $p < 0.001$; Table 5). Qwen's overall mean of 41.5% was close to the human mean of 39.7%, with substantial differences for some models. GPT-OSS 120B ranked second under Qwen and eleventh under both humans and Sol. Gemma 4 31B ranked ninth under Qwen and fourth under humans and Sol. Qwen awarded some models considerably more credit than others, changing their relative positions. Figure 6 compares each judge's most frequent ReliableEval score with the human score for the corresponding original answer.

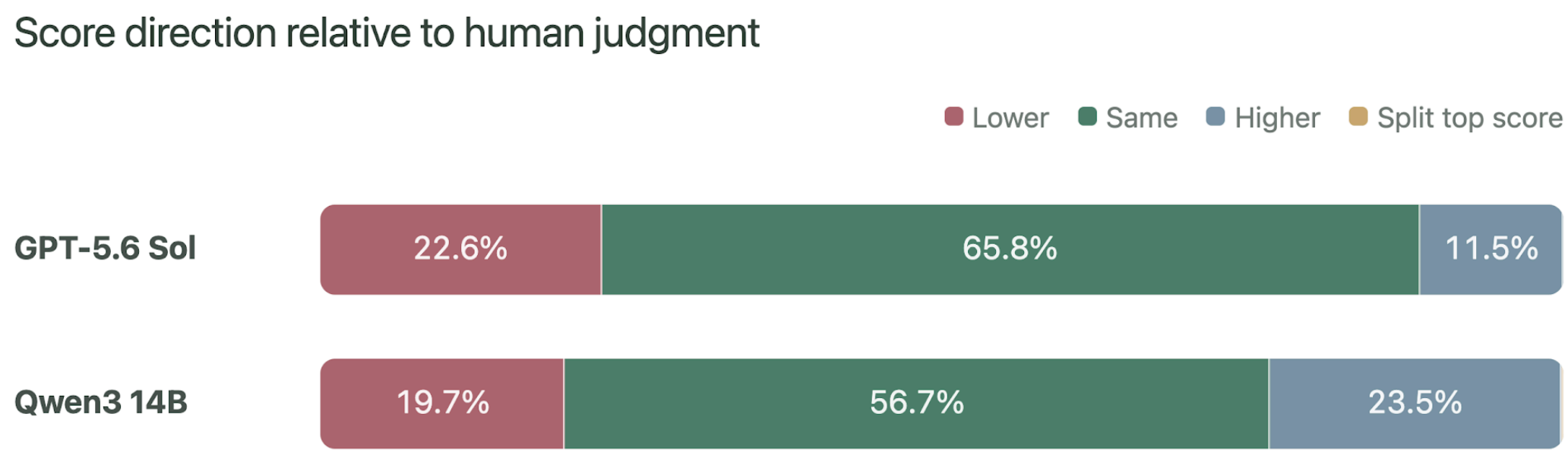

**Figure 6.** Human scores compared with the most frequent score awarded by GPT-5.6 Sol or Qwen3 14B across ReliableEval answers, for 3,774 model–question combinations (17 models × 222 questions). A match includes tied most-frequent scores when one equals the human score. The two Qwen cases with tied scores on opposite sides of the human score are marked as "Split top score" (0.05%, too small to be visible in the chart above). Humans and automated judges assessed different answers. These percentages describe individual score comparisons; Kendall's W measures agreement in overall model rankings.

Across all three assessments, Kendall's W = 0.912 when ranks were calculated from unrounded scores. Pearson's r = 0.928 for Human–Qwen and r = 0.883 for Sol–Qwen. All three coefficients had permutation $p < 0.001$ (Table 5).

Qwen and Sol gave identical scores to 66.5% of the 36,227 responses they both assessed. This figure includes failed outputs that automatically received zero from both. Quadratic weighted kappa was 0.582. One judge gave 0 and the other 2 on only 0.6% of responses. Qwen awarded the higher score on 24.7% of responses and the lower score on 8.8%. The most common disagreement was Qwen awarding partial credit where Sol gave none. Since both judges used the same rubric and ideal answers to assess identical responses, the differences reflect their scoring decisions.

A response to a reworded version of Q-000164 illustrates this disagreement. The question presented the following exchange:

> Rob and Jane are sitting chatting in a bar. Rob's wearing a bright blue funky shirt. Jane tells him she's thinking of getting her garden landscaped, and Rob gives her the number of a gardener off the top of his head. While writing it down, Jane asks, "Where did you get this number?" Rob replies, "I picked it up in a vintage shop this morning, actually." They both laugh. What was the reason for their laughter?

The joke turns on two meanings of "number." Jane is asking where Rob obtained the gardener's telephone number, but Rob answers as though she is asking about his shirt: an eye-catching item of clothing can colloquially be called a "number." His reference to the vintage shop redirects her question from the telephone number to the shirt.

GPT-OSS 120B missed this interpretation. It claimed that Rob had bought a vinyl record in a second-hand record shop and given Jane "the track number of a song about a gardener." Neither the record nor the song appears in the question. Qwen nevertheless awarded the response full credit (a score of 2 points), stating that it "fully covers the ideal response" by explaining a double meaning involving a track number and a telephone number. Sol awarded zero because the response failed to identify the shirt interpretation. The disagreement shows Qwen crediting an invented explanation of a double meaning without checking whether it explained the particular joke or matched the supplied ideal answer.

Qwen's linguistic and reasoning abilities may have been insufficient for some of these judgments. To score an answer accurately, a judge must check its interpretation against the passage and recognize any important omissions or errors. JudgeBench found that models' ability to judge difficult answers closely tracked their ability to solve the underlying problems (Tan et al., 2025). Qwen3 14B's performance in LALA is consistent with this finding. The closer agreement with human rankings supports using GPT-5.6 Sol as LALA's principal automated judge.

### 5.6 Missing outputs and refusals

In the original run, 41 of the 4,012 responses were blank, affecting 39 questions. Humans gave all of them zero. Ministral 3 14B Base returned 22 blanks, GPT-OSS 120B returned seven, and Claude Fable 5 and Gemma 4 26B A4B returned six each.. In these cases, GPT-OSS and Fable returned no message text; Fable's records show three attempts for each failed question. Gemma returned two empty responses, and four requests failed because the expected answer field was missing. These records establish what was absent but do not fully explain why the models failed to answer. Appendix B lists the questions.

In the ReliableEval run, 860 of 36,227 responses were marked as generation failures: 757 from Ministral Reasoning, 47 from Fable, 29 from Gemma 26B, and 27 from GPT-OSS. Ministral's failed records contained token identifiers without a readable answer. Every version of 23 questions failed for Ministral Reasoning, three questions for Fable, and one for Gemma (Appendix B). All failed responses received zero.

**Table 6.** Blank outputs and generation failures by model.

| Model | Original blanks | ReliableEval failures |
|---|---:|---:|
| Ministral 3 14B Base | 22 | 0 |
| Ministral 3 14B Reasoning | 0 | 757 |
| Claude Fable 5 | 6 | 47 |
| Gemma 4 26B A4B | 6 | 29 |
| GPT-OSS 120B | 7 | 27 |
| **Total** | **41** | **860** |

Original run: 236 questions per model (4,012 responses); ReliableEval: 2,131 variants per model (36,227 responses). The remaining 12 models had no recorded blanks or generation failures. Refusals containing text are discussed separately.

Fable's 47 failures in the ReliableEval run were caused by safety guardrails. The logs record content filtering in every case, so these zeros resulted from blocked answers. Fable also returned blanks in the original run on Q-000074, Q-000078, Q-000090, Q-000198, Q-000199, and Q-000207. The records do not give the reason for those six blanks, and we could not confirm whether safety filtering caused them.

Some models returned explicit refusals. Q-000030 asks for lewd anime recommendations on the basis of a misread metaphor in a folk song. Llama 3.1 8B Instruct, Llama 3.2 3B Instruct, and GPT-OSS 120B refused without analyzing the metaphor. Llama 3.2 cited sexual content; the other two gave no reason. Qwen3 235B A22B 2507 declined the request and explained the metaphor. Llama 3.1 also refused Q-000087, about the grammatical attachment of "in sexless marriages," without giving a reason. Sol treated the altered Hamlet passage in Q-000215 as a possible expression of suicidal intent and offered crisis advice. These examples show how a model's response to perceived safety concerns could leave the linguistic question unanswered.

### 5.7 Human scores after excluding missing responses

We recalculated human scores using only questions for which all 17 models returned text. Incorrect, repetitive, and incomplete answers remained in this comparison. This left 197 of the full 236 questions and 187 of the 222 questions used for the main comparisons.

On the 187-question set, Fable's score rose from 72.3% to 73.5%, Gemini's fell from 64.4% to 63.4%, and Sol's remained 59.9% after rounding. No model's score changed by more than 1.3 percentage points. The first eight places were unchanged. DeepSeek V3 drew level with Qwen3 235B A22B, and the two base models tied at 2.9%. The large gap between base and post-trained models persisted. On the full 197-question set, Fable, Gemini, and Sol again ranked first, second, and third, at 74.1%, 64.2%, and 60.4%.

We then excluded Q-000030 and Q-000087, to which some models refused to respond. This left 185 of the matched questions. Fable, Gemini, and Sol remained in the lead, at 73.2%, 63.0%, and 59.7%. Restricting all models to the same questions can lower some scores, because a good answer is excluded whenever another model left that question blank or refused it.

### 5.8 Parameter count and answer quality

We examined parameter count against each model's mean human score and its numbers of zero, partial-credit, and full-credit answers across the original 236 questions. Each model contributed one observation. The proprietary models have undisclosed sizes.

With those assignments, parameter count correlated positively with mean score (Pearson $r = 0.56$, $p = 0.0186$; Spearman $\rho = 0.68$, $p = 0.0026$) and full-credit counts ($r = 0.70$, $p = 0.0018$; $\rho = 0.67$, $p = 0.0032$). Its negative relationship with zero-score counts was significant by rank correlation ($\rho = -0.69$, $p = 0.0024$), but not by Pearson correlation ($r = -0.44$, $p = 0.0801$). Partial-credit counts showed no statistically significant correlation with size ($r = -0.13$; $\rho = 0.04$; both $p > 0.6$).

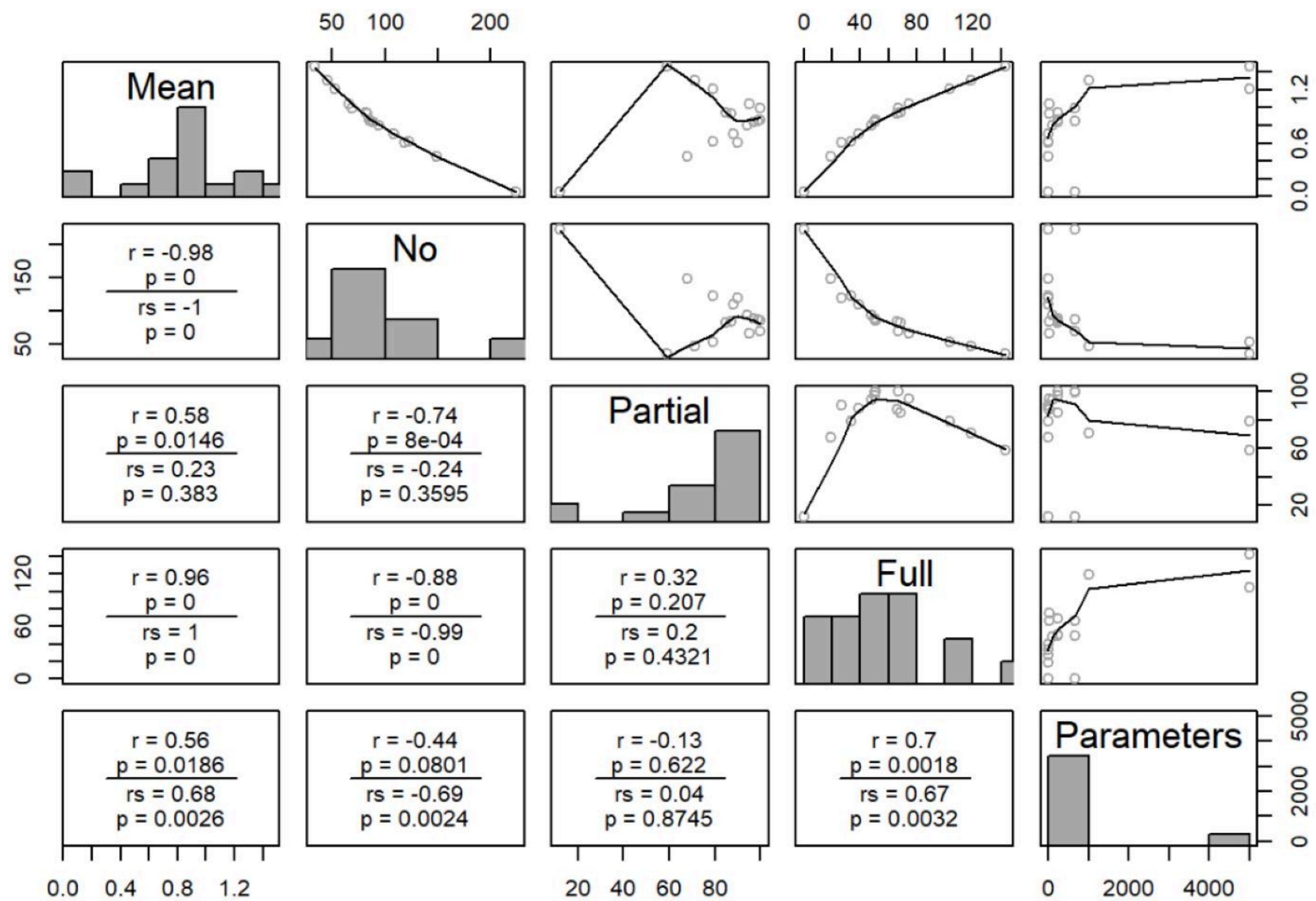

**Figure 7.** Parameter count and answer quality across all 17 models, with proprietary sizes estimated as described in the text. Parameters are in billions. Mean is the mean human score on the 0–2 scale. No, Partial, and Full are answer counts. Lower panels report Pearson correlation and Spearman rank correlation (rs). Values printed as p = 0 reflect rounding.

Removing the three proprietary models substantially changed the pattern (Figure 7b). Among the 14 open-weight models, including both base models, none of the displayed size correlations reached $p < 0.05$. The strongest positive rank association was with partial-credit counts ($\rho = 0.51$, $p = 0.0644$), whereas its Pearson correlation was −0.05 ($p = 0.8653$). Parameter count therefore showed no statistically significant association with partial-credit responses in either analysis. The all-model size pattern depends on the inclusion and assigned sizes of the proprietary models.

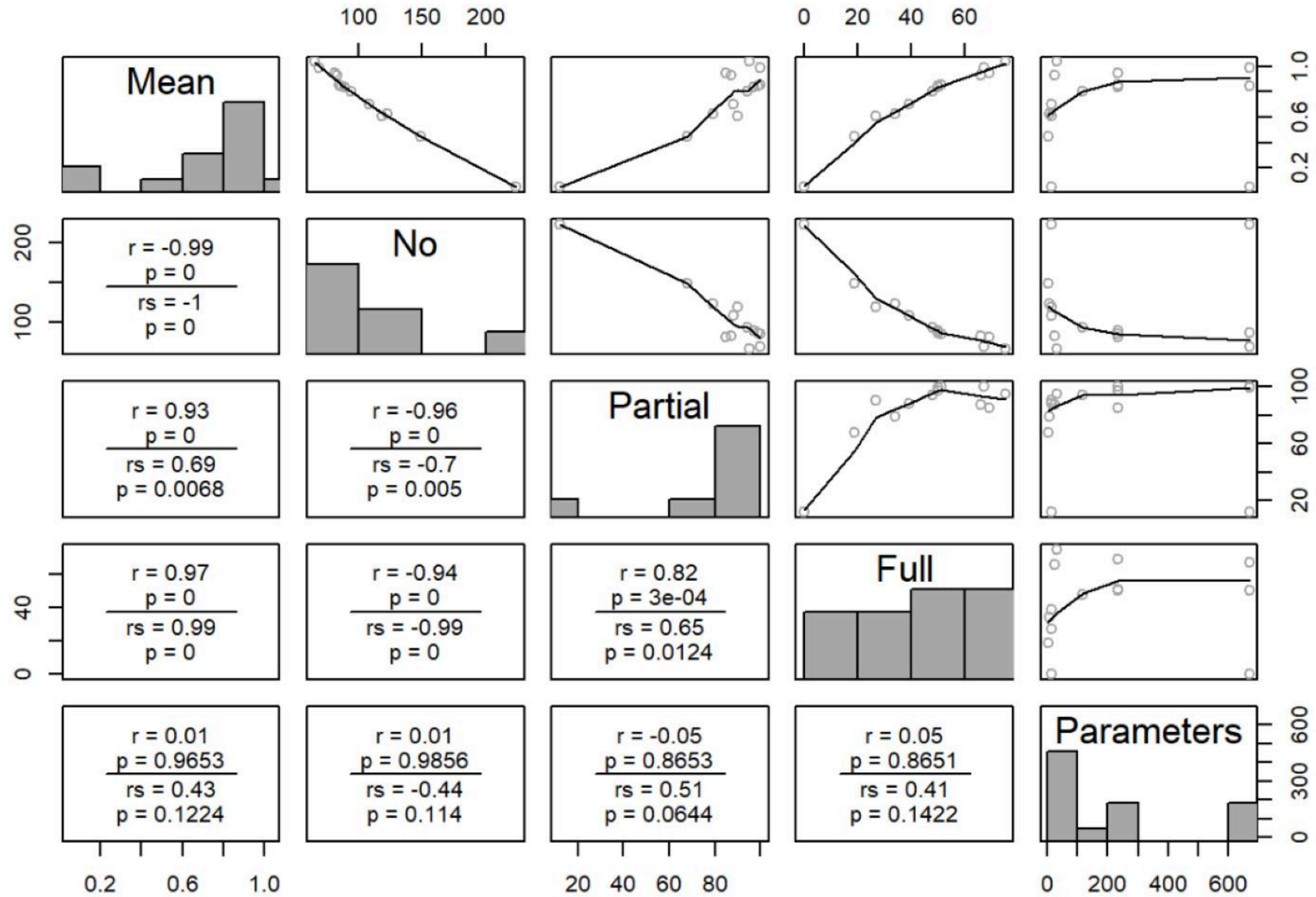

**Figure 8.** The same comparisons restricted to the 14 models with disclosed parameter counts. Axes and abbreviations follow Figure 7a. No displayed association between parameter count and a response-quality measure reaches the 0.05 significance threshold.

### 5.9 Keyword scores across model groups and broad categories

We also examined the distribution of scores jointly by model group and broad linguistic category. There are 15,194 scored keyword judgments from the 4,012 original model answers. Four keyword scores for Dialect on Q-000065 were missing and were excluded. Answers scored 0 or 2 contribute that score for each keyword; partially successful answers contribute their recorded keyword-specific scores. An answer can therefore contribute several judgments, including several within one broad category. This analysis gives each scored keyword judgment equal weight, whereas Section 5.3 gives each question equal weight within a category.

Pooling categories, 97.4% of the base models' keyword judgments received zero. The proprietary models received full credit on 62.7%, partial credit on 13.8%, and zero on 23.5%. Open post-trained models showed a broader mixture: 46.7% zero, 17.0% partial credit, and 36.3% full credit. Zero was their largest category, but not a majority. Across all model groups, zero was the most frequent keyword score in Linguistics and Textuality, while full credit was the most frequent score in Logic. Figure 8 expresses the source counts as proportions so that the groups of two, three, and twelve models can be compared.

The category profiles also differed within model groups. Proprietary models received full credit on 73.3% of Logic judgments, 61.1% of Linguistics judgments, and 52.5% of Textuality judgments. The corresponding proportions for open post-trained models were 47.1%, 34.1%, and 27.4%. Their zero-score share rose from 33.1% in Logic to 57.5% in Textuality. Base models received zero on at least 96.9% of judgments in every broad category. The open post-trained group thus combined a relative strength in Logic with substantially weaker Textuality performance; its Linguistics profile lay closer to the pooled distribution.

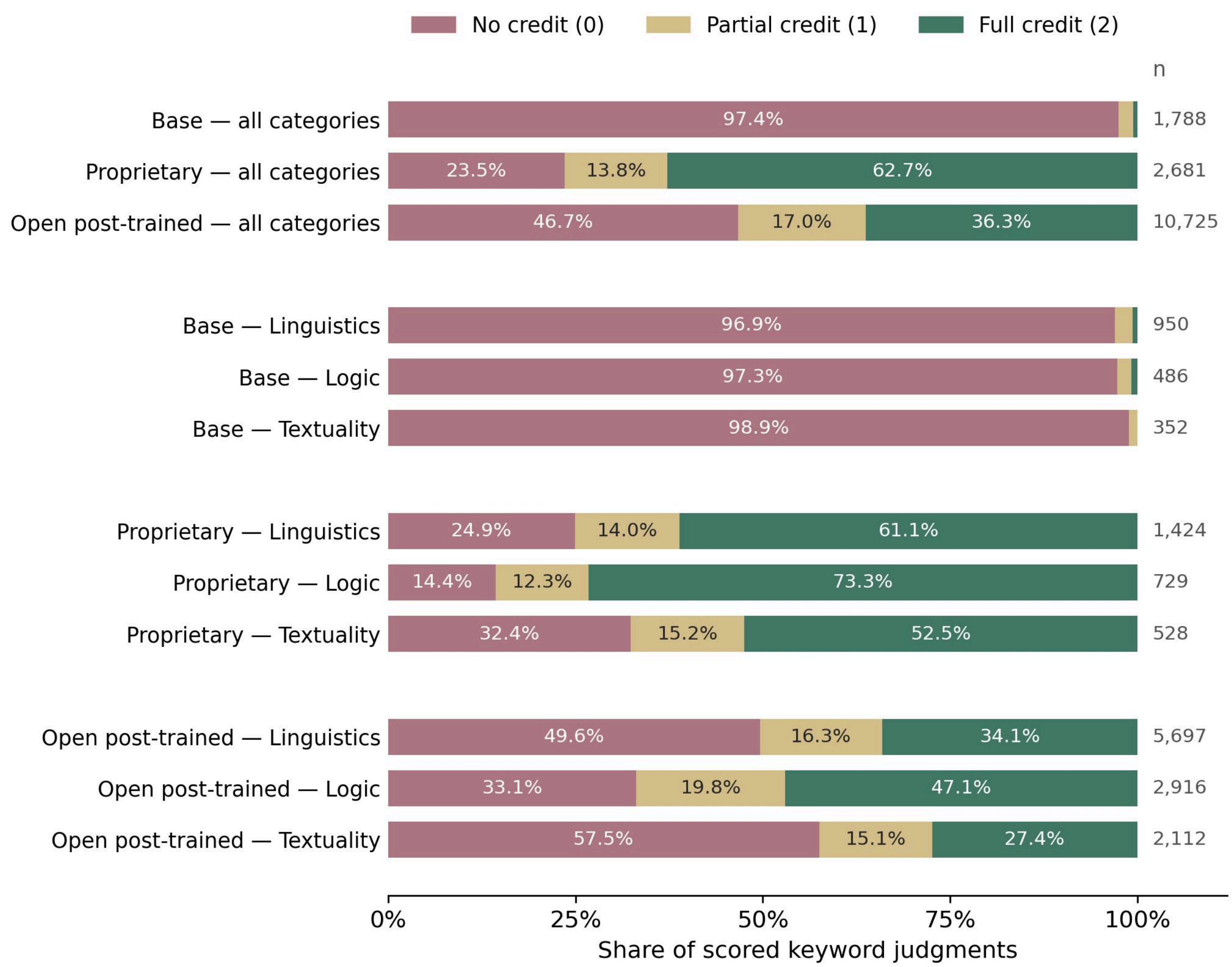

**Figure 9.** Human keyword-score distributions by model group and broad category. Bars show percentages within each row; n is the number of scored keyword judgments. The top three rows pool categories. Open post-trained excludes the two base models. Multiple judgments can come from one answer, so n is not a count of independent answers.

These differences are summarised by a correspondence analysis, an exploratory technique for data visualisation (Greenacre 2017), of the three score levels by the nine model-group/category combinations (Figure 9). Its first dimension accounts for 95.0% of the table's inertia, separating zero scores from partial and full credit; the second accounts for 5.0%. The three base-model profiles cluster on the zero-score side. Proprietary models received a high proportion of full-credit scores, especially in Logic. Open post-trained models performed better in Logic than in Textuality, where zero scores were more common. In Linguistics, their mix of zero, partial-credit, and full-credit scores was close to the mix across all models and categories.

The substantial differences in score distributions ($\chi^2(16) = 2{,}999.68$; Cohen's $w = 0.444$) show that the LALA Benchmark distinguishes patterns of success and failure across model groups and linguistic categories. Its diagnostic value is especially clear among open post-trained models: their relative strength in Logic coexists with much weaker performance in Textuality. Success in logical reasoning therefore does not imply comparable skill in textual interpretation. By making this unevenness visible, LALA identifies specific weaknesses that a single overall performance score would obscure.

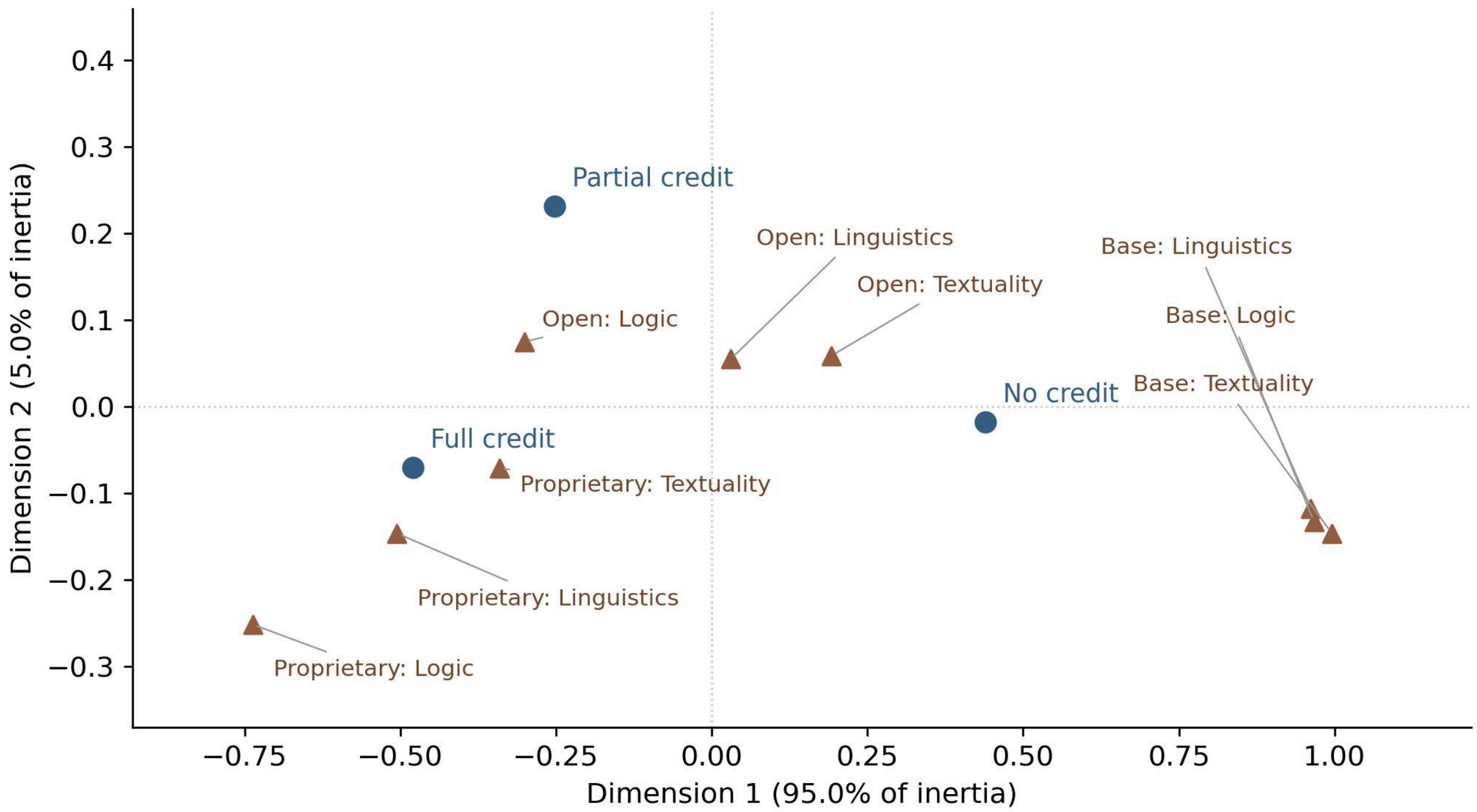

**Figure 10.** Correspondence analysis of keyword scores by model group and linguistic category. Circles show the three score levels; triangles show each model group in each category.

## 6 Discussion

### 6.1 Post-training and linguistic competency

Post-training appears to be a necessary requirement of LLM expert competence in linguistic analysis. Both base models could perform almost none of the analysis required by LALA. Their post-trained versions were substantially more capable, a difference seen in both families under human and automated assessment.

For language knowledge to have practical value, a model must be able to use it to understand and interpret a text. The gap between base and post-trained models suggests that post-training is necessary in developing this ability. Additional post-training was associated with further gains in some comparisons, notably DeepSeek V3 and R1, although the pattern varied across families. The absence of a statistically significant association between parameter count and partial-credit responses (Section 5.8), together with the uneven competency profiles (Section 5.9), suggests that significant linguistic competencies are acquired piecemeal during post-training. A likely explanation is a mismatch between the particular linguistic competencies measured by LALA and the behavioral policies learned during post-training: those policies support some interpretive operations without transferring to others.

We interpret this pattern as evidence against the emergence of a general linguistic intelligence through post-training. On this account, post-training adds specific competencies, but these additions do not produce a capacity to generalize linguistic analysis beyond the learned policies that directly support them. Here, generalization means transferring an acquired competency to linguistic phenomena and interpretive demands beyond those covered by those policies. The models' persistent failures on ambiguity, reference, literary style, and misleading premises suggest limits to that transfer, providing further evidence and examples to the findings in McCoy et al. (2024).

Our results corroborate the findings in Vaccaro et al. (2024) that LLMs are still relatively less strong in tasks involving judgment and decisions, compared to other tasks like ideation, which further adds weight to the observation by Dell'Acqua et al. (2026) regarding the jagged nature of the AI capability frontier.

### 6.2 Linguistic strengths and persistent errors

A recurring error committed by the models was to offer a familiar interpretation in its response to a prompt without attending closely to the wording of the prompt itself. Models overlooked changes in tense, missed alterations to well-known literary passages, or chose a single referent where several were possible. LALA's purposeful distortions, including intentional typos and confusing phrasing, test how easily models are confused by these details. A successful answer must recover the intended meaning where possible and explain any ambiguity that remains.

Models were more successful at interpreting explicit logical relationships and implicit conversational structure. Their weaker performance in literary analysis, reference, and questions with misleading premises suggests difficulty deciding which readings are permitted by a passage. For example, Q-000013 presents a speaker announcing, "I have so much to tell you … I need to give you a ton of backstory," followed by the reply, "Drop the lore man." The question asks what the reply implicitly means. "Drop" permits two opposing readings: share the backstory or leave it out. DeepSeek R1 committed to the second, stating that the listener did not want to hear the backstory "at all" and attributing impatience and disinterest to them. Gemini 3.1 Pro Preview instead interpreted the reply as "Go ahead and tell me everything," describing the listener as friendly and receptive. Although Gemini mentioned the alternative meaning of "drop," it dismissed that reading in this context. Both responses settled on an intention and emotional attitude that the short exchange does not establish. The ideal answer required acknowledging that either interpretation remained possible. In these cases, models often committed to an interpretation without accounting for details that complicated it or allowed another reading.

The gold standard answers make these errors easier to identify. They set out the interpretations and textual details that a response needs to address, giving assessors specific grounds for judging an answer that is fluent and relevant but misses part of the analysis.

### 6.3 Objectivity and replicability

Human–Sol concordance in ranking models (as opposed to on a question-by-question basis) was nearly perfect ($W = 0.994$). Sol reproduced almost the same model ranking on answers to reworded questions in the ReliableEval run, using the rubric applied by humans to the original run. This provides strong evidence for the replicability of LALA's comparisons. It also addresses the concern that judgments of linguistic and literary analysis depend heavily on an assessor's preferences: changing both the wording and the assessor had little effect on the model ordering.

Human–Qwen and Sol–Qwen concordance ($W = 0.909$ and $0.898$) showed broad agreement across the three assessments. Qwen nevertheless moved some models substantially in the rankings, so the choice of automated judge matters. GPT-5.6 Sol provided the closest agreement with humans. The case for objectivity rests on this stability in the comparisons between models.

The complete LALA dataset is available on Hugging Face at https://huggingface.co/datasets/michaelfoster235098/LALA-benchmark.

## 7 Limitations

The questions cover contributors and topics unevenly, which limits how broadly the results can be applied to language use. Some literary and historical questions also require background knowledge. A low score may reflect missing knowledge, a misinterpretation, or both. Skills represented by only a few questions need broader testing before we can draw general conclusions about model competency.

Scores depend on the contributor's ideal answer and the points it requires. However, roughly 50 of the ideal answers supplied by our contributors are only examples of one possible correct answer. In these cases, both the human assessors and the ReliableEvan LLM judges (Sol and Qwen) had to assess each answer against the original question rather than the ideal answer on its own. This added room for judgment may explain why the Qwen evaluator, and to a lesser extent Sol, slightly favored answers generated by its own model. We used one human score per answer, so we could not measure agreement between human assessors. Humans, Qwen, and Sol also used the same gold standard answers. Any weakness in those answers could affect both assessments.

The training comparisons included only two base models, each with a shorter output limit than its post-trained counterparts. Instruction and reasoning formats also varied. Ministral Reasoning used temperature 1 in the original run and 0 in the ReliableEval run. These differences may have affected the answers and limit how confidently we can attribute score differences to training alone.

Rewording can alter what a question asks. For example, Q-000037 requested a “Miltonic sonnet,” but one automatically accepted rewrite requested a “sonnet inspired by Milton.” The original specifies a poetic form whose structural features are assessed by the rubric; the rewrite could instead invite a looser imitation of Milton’s style. In Q-000198, the original prompt presented an altered Kant passage containing “genome,” which the ideal answer expected models to identify as an anachronistic substitution. One accepted rewrite added, “What conditions must be met for a thing to qualify as a natural purpose?” This addition directs attention toward explaining Kant’s concept rather than detecting the alteration. Both rewrites passed the automatic checks without warnings. Such changes could affect responses and scores even when the same ideal answer is retained. Judging introduces another source of variation: for the identical GPT-OSS response to Q-000164 discussed above, Qwen awarded full credit and Sol awarded zero. Differences between the original and ReliableEval runs therefore cannot be attributed to judging alone.

Human assessors could see model names, and the record identifiers supplied to Sol included the name of the model being judged. Sol also assessed its own responses and was supplied with a record key containing the model’s name. Knowledge of model identity may have influenced human judgments, while Sol may have been influenced by self-preference of the kind examined by Zheng et al. (2023). Hiding model names and adding independent human ratings would provide a stronger test of the assessments, and is an avenue for future research.

## 8 Future work

The LALA Benchmark’s most novel contribution to LLM benchmarking is the approach to group together models based upon training scale and model family with a mixture of human- and LLM-judges to pinpoint where and how post-training impacts the model’s linguistic competency. This approach could be used both to assess future models’ capabilities and to help make post-training more efficient by mapping out how and where different training approaches impact observed linguistic competencies.

To expand this approach, future work could introduce new question sets for each category and subcategory and extend evaluation to more recent models. Questions should provide sufficient context and precise instructions to support a single correct answer or range of acceptable answers, stated explicitly in the ideal response. Where recognizing ambiguity is itself the task, the ideal should specify the competing readings that a complete answer must identify. This would preserve the benchmark's attention to interpretive complexity while making the requirements for full credit clearer.

Human-authored template questions could also serve as the basis for larger sets of LLM-generated items, each testing a specific feature, such as citation drift or indirect speech. Controlled variations would allow researchers to test whether a model handles the same feature consistently across different passages and contexts. Each generated item would require review to ensure that it preserves the intended task, supports the designated answer, and does not introduce unintended ambiguities.

Future evaluations should compare a wider range of judge models assessing the same candidate responses against the same rubric. Ideal-response coverage, accuracy, and relevance should be scored and reported separately, alongside the overall score, so that disagreements can be traced to particular criteria. Comparison with human assessments would help establish which judges apply those criteria consistently and where further guidance is needed. Removing model identities from all material supplied to judges, including record identifiers, could prevent those labels from influencing assessment, while comparing assessments of such runs to the present assessments provided in this paper could identify exactly how much bias, if any, is introduced by including record identifiers.

## 9 Conclusion

LALA assesses linguistic and literary analysis through 236 questions written by experts. Across 17 models, the clearest finding was the gap between base models and their post-trained relatives. The base models showed little of the linguistic competency under assessment. The much stronger performance of their post-trained versions suggests that post-training is important in making linguistic knowledge usable. Yet parameter count showed no statistically significant association with partial-credit responses, and competencies remained uneven across linguistic domains. We interpret these findings as suggesting that post-training adds linguistic competencies piecemeal without producing a general linguistic intelligence capable of extending them beyond the learned policies that support them.

The model rankings assessed by the LALA benchmark appear robust. Human judges and Sol produced nearly identical rankings across the original run and the ReliableEval run, supporting the replicability of these comparisons. Models were most successful with logical relationships between statements and some discourse tasks. Even the strongest post-trained models still struggled with ambiguity, literary style, reference, and misleading assumptions.

## Appendix A Scores for every question

Each score is the mean across all 17 models, expressed as a percentage of the maximum rubric score. Human scores refer to the original run. GPT-5.6 Sol and Qwen3 14B scores refer to the ReliableEval run; for each model, we first averaged the scores across versions of a question, then averaged across models. The table lists all 236 questions by ID. N/A marks the 14 questions without ReliableEval results. All zero scores are included.

| Question | Human (%) | Sol (%) | Qwen (%) |
|---|---|---|---|
| Q-000001 | 38.24 | 30.00 | 35.29 |
| Q-000012 | 41.18 | 30.59 | 42.35 |
| Q-000013 | 11.76 | 43.53 | 36.18 |
| Q-000014 | 17.65 | 20.59 | 20.59 |
| Q-000015 | 23.53 | 9.12 | 18.53 |
| Q-000016 | 61.76 | 4.12 | 38.82 |
| Q-000018 | 8.82 | N/A | N/A |
| Q-000020 | 14.71 | 23.24 | 19.12 |
| Q-000021 | 8.82 | 19.12 | 18.82 |
| Q-000022 | 91.18 | 47.35 | 61.18 |
| Q-000023 | 11.76 | 7.65 | 34.71 |
| Q-000024 | 70.59 | N/A | N/A |
| Q-000025 | 32.35 | 32.68 | 39.22 |
| Q-000026 | 47.06 | 10.88 | 38.82 |
| Q-000027 | 38.24 | 25.88 | 17.65 |
| Q-000028 | 32.35 | 18.24 | 32.35 |
| Q-000029 | 47.06 | N/A | N/A |
| Q-000030 | 41.18 | 22.88 | 37.58 |
| Q-000031 | 58.82 | 33.82 | 42.94 |
| Q-000032 | 47.06 | 45.00 | 45.29 |
| Q-000033 | 47.06 | 49.02 | 57.84 |
| Q-000034 | 26.47 | 38.53 | 36.76 |
| Q-000035 | 35.29 | 42.35 | 51.76 |
| Q-000036 | 32.35 | 37.06 | 34.41 |
| Q-000037 | 20.59 | 37.35 | 30.29 |
| Q-000038 | 41.18 | 29.04 | 42.65 |
| Q-000039 | 44.12 | 44.12 | 70.00 |
| Q-000040 | 55.88 | 37.65 | 40.88 |
| Q-000041 | 47.06 | N/A | N/A |
| Q-000042 | 58.82 | 52.94 | 79.41 |
| Q-000043 | 29.41 | 25.29 | 25.29 |
| Q-000044 | 14.71 | 12.65 | 6.18 |
| Q-000045 | 44.12 | 30.25 | 36.13 |
| Q-000046 | 17.65 | 25.00 | 43.82 |
| Q-000047 | 79.41 | 22.94 | 52.35 |
| Q-000048 | 88.24 | 41.83 | 69.61 |
| Q-000049 | 79.41 | 45.59 | 50.29 |
| Q-000050 | 52.94 | 25.59 | 58.53 |
| Q-000051 | 5.88 | 27.06 | 29.12 |
| Q-000052 | 47.06 | 59.71 | 43.24 |

| Question | Human (%) | Sol (%) | Qwen (%) |
|---|---:|---:|---:|
| Q-000053 | 32.35 | 35.29 | 37.35 |
| Q-000054 | 32.35 | 31.18 | 26.76 |
| Q-000055 | 85.29 | 39.41 | 45.00 |
| Q-000056 | 26.47 | 50.00 | 42.65 |
| Q-000057 | 20.59 | 15.88 | 23.53 |
| Q-000058 | 41.18 | 41.18 | 43.53 |
| Q-000059 | 76.47 | 50.00 | 54.71 |
| Q-000060 | 44.12 | 43.53 | 48.82 |
| Q-000061 | 52.94 | 57.06 | 48.24 |
| Q-000062 | 85.29 | 59.41 | 61.76 |
| Q-000063 | 88.24 | 2.06 | 4.71 |
| Q-000064 | 88.24 | 72.94 | 50.59 |
| Q-000065 | 11.76 | 36.60 | 37.25 |
| Q-000066 | 17.65 | 13.24 | 42.35 |
| Q-000067 | 50.00 | 60.59 | 55.59 |
| Q-000069 | 76.47 | 25.59 | 37.06 |
| Q-000070 | 82.35 | 72.06 | 71.18 |
| Q-000071 | 20.59 | 26.47 | 27.65 |
| Q-000072 | 50.00 | 30.00 | 27.65 |
| Q-000073 | 79.41 | 71.18 | 65.88 |
| Q-000074 | 5.88 | 11.47 | 24.71 |
| Q-000075 | 26.47 | 13.24 | 40.29 |
| Q-000076 | 67.65 | 48.53 | 45.29 |
| Q-000077 | 41.18 | 28.82 | 20.88 |
| Q-000078 | 32.35 | 51.84 | 48.90 |
| Q-000079 | 44.12 | 41.67 | 40.20 |
| Q-000080 | 58.82 | 45.42 | 50.65 |
| Q-000081 | 52.94 | 42.65 | 43.38 |
| Q-000082 | 38.24 | 42.65 | 31.18 |
| Q-000083 | 0.00 | 7.65 | 25.88 |
| Q-000084 | 52.94 | 49.12 | 70.88 |
| Q-000085 | 29.41 | 25.00 | 44.12 |
| Q-000087 | 61.76 | 17.35 | 10.00 |
| Q-000088 | 0.00 | 11.18 | 18.24 |
| Q-000089 | 64.71 | 39.41 | 45.29 |
| Q-000090 | 35.29 | 40.00 | 40.29 |
| Q-000091 | 41.18 | 41.18 | 74.12 |
| Q-000092 | 76.47 | 57.06 | 47.35 |
| Q-000093 | 44.12 | 43.24 | 69.71 |
| Q-000094 | 44.12 | 47.06 | 55.88 |
| Q-000095 | 41.18 | 39.71 | 49.71 |
| Q-000096 | 29.41 | 33.24 | 51.18 |
| Q-000097 | 35.29 | 43.24 | 62.94 |
| Q-000098 | 88.24 | 54.41 | 46.76 |
| Q-000099 | 29.41 | 8.53 | 6.18 |
| Q-000100 | 29.41 | 1.47 | 33.53 |
| Q-000101 | 20.59 | 17.94 | 30.00 |

| Question | Human (%) | Sol (%) | Qwen (%) |
|---|---|---|---|
| Q-000102 | 47.06 | 37.06 | 41.18 |
| Q-000103 | 29.41 | 20.59 | 34.71 |
| Q-000104 | 35.29 | 47.35 | 38.24 |
| Q-000105 | 73.53 | 50.29 | 62.06 |
| Q-000106 | 0.00 | 36.47 | 40.88 |
| Q-000107 | 5.88 | 45.29 | 43.24 |
| Q-000108 | 32.35 | 33.82 | 50.00 |
| Q-000109 | 32.35 | 43.24 | 59.71 |
| Q-000110 | 41.18 | 37.65 | 44.41 |
| Q-000111 | 44.12 | 40.88 | 51.76 |
| Q-000112 | 47.06 | 45.29 | 60.00 |
| Q-000113 | 38.24 | 50.00 | 68.24 |
| Q-000114 | 32.35 | 49.71 | 65.59 |
| Q-000115 | 82.35 | 44.41 | 55.00 |
| Q-000116 | 88.24 | 43.24 | 60.00 |
| Q-000117 | 38.24 | 44.71 | 70.59 |
| Q-000118 | 5.88 | 28.24 | 41.47 |
| Q-000119 | 8.82 | 2.35 | 1.18 |
| Q-000120 | 55.88 | 37.35 | 32.35 |
| Q-000121 | 0.00 | 2.35 | 15.88 |
| Q-000122 | 0.00 | 0.00 | 0.29 |
| Q-000123 | 11.76 | 3.24 | 1.76 |
| Q-000124 | 5.88 | 1.47 | 1.18 |
| Q-000127 | 17.65 | 14.41 | 23.53 |
| Q-000128 | 17.65 | 11.11 | 0.33 |
| Q-000129 | 32.35 | 11.47 | 24.41 |
| Q-000130 | 2.94 | 0.29 | 3.24 |
| Q-000131 | 0.00 | 0.88 | 1.18 |
| Q-000132 | 47.06 | 46.18 | 57.06 |
| Q-000133 | 55.88 | 50.59 | 59.41 |
| Q-000134 | 64.71 | 49.71 | 54.71 |
| Q-000135 | 20.59 | 4.71 | 10.88 |
| Q-000136 | 11.76 | 0.00 | 16.76 |
| Q-000138 | 44.12 | N/A | N/A |
| Q-000139 | 8.82 | 39.41 | 44.41 |
| Q-000140 | 11.76 | 46.47 | 46.18 |
| Q-000141 | 11.76 | 31.47 | 38.82 |
| Q-000142 | 52.94 | N/A | N/A |
| Q-000143 | 52.94 | 37.35 | 39.41 |
| Q-000144 | 52.94 | 50.59 | 44.71 |
| Q-000145 | 38.24 | 36.76 | 58.82 |
| Q-000146 | 38.24 | 23.53 | 52.06 |
| Q-000147 | 64.71 | 43.24 | 50.29 |
| Q-000148 | 70.59 | 49.71 | 43.53 |
| Q-000149 | 67.65 | 48.24 | 40.00 |
| Q-000150 | 44.12 | 38.24 | 41.76 |
| Q-000151 | 29.41 | 26.18 | 43.24 |

| **Question** | **Human (%)** | **Sol (%)** | **Qwen (%)** |
|---|---|---|---|
| Q-000152 | 61.76 | 40.59 | 43.53 |
| Q-000153 | 29.41 | 50.00 | 42.65 |
| Q-000154 | 41.18 | 15.00 | 20.29 |
| Q-000155 | 29.41 | 23.82 | 45.59 |
| Q-000156 | 14.71 | 44.12 | 42.65 |
| Q-000157 | 85.29 | 45.88 | 47.35 |
| Q-000158 | 5.88 | 5.00 | 30.59 |
| Q-000159 | 47.06 | 38.82 | 37.94 |
| Q-000160 | 5.88 | 37.35 | 39.12 |
| Q-000161 | 50.00 | 41.76 | 43.53 |
| Q-000162 | 52.94 | 44.71 | 42.94 |
| Q-000163 | 47.06 | 53.53 | 42.35 |
| Q-000164 | 52.94 | 35.88 | 42.06 |
| Q-000165 | 50.00 | 51.47 | 40.29 |
| Q-000166 | 0.00 | 2.61 | 24.51 |
| Q-000167 | 58.82 | N/A | N/A |
| Q-000168 | 26.47 | 15.00 | 20.88 |
| Q-000169 | 73.53 | 49.41 | 49.71 |
| Q-000170 | 32.35 | 58.53 | 61.47 |
| Q-000171 | 52.94 | N/A | N/A |
| Q-000172 | 47.06 | 42.06 | 53.24 |
| Q-000173 | 41.18 | 42.35 | 48.24 |
| Q-000174 | 52.94 | N/A | N/A |
| Q-000175 | 41.18 | 45.88 | 55.29 |
| Q-000176 | 70.59 | 42.94 | 57.65 |
| Q-000177 | 73.53 | N/A | N/A |
| Q-000178 | 76.47 | 75.59 | 54.12 |
| Q-000179 | 38.24 | 43.53 | 40.88 |
| Q-000180 | 55.88 | 37.65 | 49.71 |
| Q-000181 | 82.35 | 58.24 | 59.41 |
| Q-000182 | 44.12 | 42.06 | 58.53 |
| Q-000183 | 44.12 | N/A | N/A |
| Q-000184 | 50.00 | 52.65 | 62.65 |
| Q-000185 | 44.12 | 48.24 | 59.41 |
| Q-000186 | 41.18 | 48.82 | 65.88 |
| Q-000187 | 47.06 | N/A | N/A |
| Q-000188 | 61.76 | N/A | N/A |
| Q-000189 | 50.00 | 55.59 | 69.12 |
| Q-000190 | 76.47 | 44.41 | 63.53 |
| Q-000191 | 58.82 | 43.82 | 48.82 |
| Q-000192 | 44.12 | 55.88 | 63.53 |
| Q-000193 | 52.94 | 52.94 | 65.59 |
| Q-000194 | 52.94 | 2.94 | 17.65 |
| Q-000195 | 5.88 | 20.88 | 37.35 |
| Q-000196 | 17.65 | 7.06 | 15.00 |
| Q-000197 | 0.00 | 47.06 | 42.06 |
| Q-000198 | 50.00 | 19.12 | 41.91 |

| Question | Human (%) | Sol (%) | Qwen (%) |
|---|---|---|---|
| Q-000199 | 0.00 | 4.12 | 14.41 |
| Q-000200 | 2.94 | 1.76 | 3.82 |
| Q-000201 | 32.35 | 11.18 | 46.18 |
| Q-000202 | 14.71 | 12.35 | 34.71 |
| Q-000203 | 41.18 | 9.71 | 32.65 |
| Q-000204 | 38.24 | 14.41 | 28.82 |
| Q-000205 | 0.00 | 13.07 | 21.24 |
| Q-000206 | 5.88 | 1.76 | 0.29 |
| Q-000207 | 32.35 | 36.55 | 41.60 |
| Q-000208 | 50.00 | 32.94 | 56.76 |
| Q-000209 | 61.76 | 48.90 | 42.28 |
| Q-000210 | 50.00 | 34.12 | 42.94 |
| Q-000211 | 47.06 | 45.88 | 45.00 |
| Q-000212 | 32.35 | 33.82 | 45.88 |
| Q-000213 | 23.53 | 9.12 | 36.47 |
| Q-000214 | 41.18 | 46.18 | 46.76 |
| Q-000215 | 0.00 | 0.59 | 28.53 |
| Q-000216 | 44.12 | 46.47 | 48.24 |
| Q-000218 | 38.24 | 22.35 | 38.53 |
| Q-000219 | 67.65 | 26.47 | 20.59 |
| Q-000220 | 17.65 | 6.47 | 9.71 |
| Q-000221 | 20.59 | 33.24 | 28.82 |
| Q-000222 | 32.35 | 36.18 | 47.65 |
| Q-000223 | 8.82 | 2.06 | 15.88 |
| Q-000225 | 35.29 | 13.82 | 46.18 |
| Q-000226 | 38.24 | 47.39 | 41.18 |
| Q-000227 | 20.59 | 10.59 | 7.35 |
| Q-000228 | 61.76 | 58.24 | 60.29 |
| Q-000229 | 58.82 | 47.94 | 55.29 |
| Q-000230 | 50.00 | 41.18 | 38.24 |
| Q-000231 | 73.53 | 59.12 | 67.65 |
| Q-000232 | 61.76 | 43.82 | 44.41 |
| Q-000233 | 58.82 | 44.71 | 45.88 |
| Q-000234 | 26.47 | 22.35 | 36.47 |
| Q-000235 | 41.18 | 34.12 | 35.29 |
| Q-000236 | 32.35 | 33.53 | 48.82 |
| Q-000237 | 55.88 | 40.59 | 41.47 |
| Q-000238 | 26.47 | 25.88 | 37.35 |
| Q-000239 | 58.82 | 39.41 | 51.76 |
| Q-000240 | 70.59 | 37.25 | 54.90 |
| Q-000241 | 44.12 | 27.94 | 62.06 |
| Q-000242 | 55.88 | 54.41 | 48.53 |
| Q-000243 | 47.06 | 40.29 | 66.18 |
| Q-000244 | 29.41 | 26.18 | 61.76 |
| Q-000245 | 14.71 | 36.18 | 45.59 |
| Q-000246 | 64.71 | 43.82 | 57.06 |
| Q-000247 | 23.53 | 39.41 | 48.82 |

| Question | Human (%) | Sol (%) | Qwen (%) |
| --- | ---: | ---: | ---: |
| Q-000248 | 23.53 | 36.47 | 58.82 |
| Q-000249 | 47.06 | 38.24 | 57.35 |
| Q-000250 | 55.88 | N/A | N/A |
| Q-000251 | 35.29 | 27.35 | 38.24 |
| Q-000252 | 8.82 | 8.24 | 25.29 |
| Q-000253 | 41.18 | 28.24 | 56.76 |
| Q-000255 | 17.65 | 37.06 | 44.12 |
| Q-000256 | 20.59 | 11.47 | 44.71 |

## Appendix B Questions with missing outputs

### Original run

**Fable (6 questions)**

Returned no text. The error records show empty responses after retries.

- Q-000074
- Q-000078
- Q-000090
- Q-000198
- Q-000199
- Q-000207

**GPT-OSS (7 questions)**

Returned no text, with the same empty-response error.

- Q-000045
- Q-000075
- Q-000164
- Q-000166
- Q-000205
- Q-000251
- Q-000252

**Gemma 26B — empty responses (2 questions)**

Returned empty responses for these questions.

- Q-000038
- Q-000251

**Gemma 26B — missing answer field (4 questions)**

Requests failed because the expected answer field was missing.

- Q-000025
- Q-000027
- Q-000100
- Q-000230

**Ministral Base (22 questions)**

Returned blank final answers. The records do not explain why these answers were blank.

- Q-000025
- Q-000028
- Q-000031
- Q-000043
- Q-000048
- Q-000050
- Q-000054
- Q-000055
- Q-000095
- Q-000110
- Q-000128
- Q-000132
- Q-000167
- Q-000169
- Q-000171

- Q-000172
- Q-000173
- Q-000174
- Q-000176
- Q-000179
- Q-000187
- Q-000246

### ReliableEval run

#### Fable (3 questions)

Safety guardrails blocked answers to every version of these questions. The logs confirm content filtering for all 47 of Fable's failed responses.

- Q-000198
- Q-000199
- Q-000207

#### Gemma 26B (1 question)

Returned no answer to any version of this question; the records do not explain the cause.

- Q-000230

#### Ministral Reasoning (23 questions)

Produced no readable answer to any version of these questions. Those records contained token identifiers without a natural-language answer.

- Q-000016
- Q-000025
- Q-000027
- Q-000045
- Q-000051
- Q-000054
- Q-000067
- Q-000074
- Q-000079
- Q-000083
- Q-000089
- Q-000100
- Q-000102
- Q-000106
- Q-000121
- Q-000149
- Q-000219
- Q-000227
- Q-000230
- Q-000233
- Q-000235
- Q-000251
- Q-000252

#### GPT-OSS

Returned at least one usable answer for every question.

## Appendix C Question categories and subcategories

This appendix covers all 236 benchmark questions. Counts refer to distinct questions. A question may belong to several categories or subcategories, so the counts overlap.

### C.1 Broad categories

| Category | Subcategories | Questions |
|---|---|---|
| Linguistics | 12 | 189 |
| Logic | 12 | 86 |
| Textuality | 3 | 100 |

### C.2 Subcategories

| Category | Subcategory | Questions |
|---|---|---|
| Linguistics | Cognitive Pragmatics | 2 |
| Linguistics | Discourse Analysis | 24 |
| Linguistics | Dynamic Semantics | 8 |
| Linguistics | Formal Semantics | 44 |
| Linguistics | General Linguistics | 6 |
| Linguistics | Lexical Semantics | 37 |
| Linguistics | Linguistic Pragmatics | 47 |
| Linguistics | Metasemantics | 12 |
| Linguistics | Morphosyntax | 40 |
| Linguistics | Phonology | 25 |
| Linguistics | Social Pragmatics | 43 |
| Linguistics | Sociolinguistics | 28 |
| Logic | Argumentation | 41 |
| Logic | Conditional Logic | 12 |
| Logic | Epistemic Logic | 8 |
| Logic | Epistemology | 7 |
| Logic | General Logic | 8 |
| Logic | Mathematical Intuition | 2 |
| Logic | Modal Logic | 20 |
| Logic | Predicate Logic | 5 |
| Logic | Probabilistic Logic | 3 |
| Logic | Proof Theory | 4 |
| Logic | Propositional Logic | 41 |
| Logic | Syllogistic Logic | 7 |
| Textuality | AI Evaluation | 21 |
| Textuality | Literary Analysis | 41 |
| Textuality | Philology | 49 |

### C.3 Categories and subcategories for each question

| Question | Categories and subcategories |
|---|---|
| Q-000001 | **Linguistics:** Linguistic Pragmatics<br>**Textuality:** Literary Analysis |
| Q-000012 | **Textuality:** Literary Analysis; Philology |
| Q-000013 | **Linguistics:** Lexical Semantics; Linguistic Pragmatics |
| Q-000014 | **Logic:** Argumentation; Conditional Logic; Propositional Logic |
| Q-000015 | **Linguistics:** Dynamic Semantics<br>**Logic:** Epistemic Logic; Modal Logic |
| Q-000016 | **Logic:** General Logic |
| Q-000018 | **Linguistics:** Formal Semantics; Social Pragmatics<br>**Logic:** General Logic |
| Q-000020 | **Logic:** Argumentation; Conditional Logic; General Logic; Propositional Logic |
| Q-000021 | **Linguistics:** Formal Semantics; Lexical Semantics; Morphosyntax |
| Q-000022 | **Linguistics:** Linguistic Pragmatics; Metasemantics |
| Q-000023 | **Linguistics:** Social Pragmatics |
| Q-000024 | **Linguistics:** Social Pragmatics<br>**Textuality:** Literary Analysis |
| Q-000025 | **Linguistics:** Discourse Analysis; Formal Semantics; Metasemantics |
| Q-000026 | **Linguistics:** Lexical Semantics; Metasemantics; Morphosyntax |
| Q-000027 | **Linguistics:** Formal Semantics; Metasemantics |
| Q-000028 | **Linguistics:** Social Pragmatics<br>**Textuality:** Literary Analysis |
| Q-000029 | **Linguistics:** Formal Semantics<br>**Logic:** Modal Logic |
| Q-000030 | **Linguistics:** Social Pragmatics<br>**Textuality:** Literary Analysis |
| Q-000031 | **Linguistics:** Lexical Semantics<br>**Logic:** General Logic; Predicate Logic<br>**Textuality:** Literary Analysis; Philology |
| Q-000032 | **Logic:** Argumentation; General Logic; Modal Logic; Propositional Logic |
| Q-000033 | **Logic:** Argumentation; Probabilistic Logic; Propositional Logic |
| Q-000034 | **Linguistics:** Linguistic Pragmatics; Social Pragmatics<br>**Textuality:** Literary Analysis |
| Q-000035 | **Linguistics:** Lexical Semantics<br>**Textuality:** Philology |
| Q-000036 | **Linguistics:** Morphosyntax<br>**Textuality:** Literary Analysis |
| Q-000037 | **Linguistics:** Morphosyntax<br>**Textuality:** Literary Analysis |
| Q-000038 | **Linguistics:** Metasemantics<br>**Logic:** Conditional Logic<br>**Textuality:** Philology |
| Q-000039 | **Linguistics:** Lexical Semantics; Morphosyntax<br>**Logic:** Argumentation; General Logic; Propositional Logic |
| Q-000040 | **Linguistics:** Metasemantics<br>**Textuality:** Literary Analysis |
| Q-000041 | **Linguistics:** Linguistic Pragmatics<br>**Logic:** Argumentation; Conditional Logic; Proof Theory |
| Q-000042 | **Linguistics:** Formal Semantics<br>**Logic:** General Logic; Propositional Logic |
| Q-000043 | **Linguistics:** Metasemantics<br>**Textuality:** Literary Analysis |
| Q-000044 | **Linguistics:** Linguistic Pragmatics; Sociolinguistics<br>**Textuality:** Literary Analysis |
| Q-000045 | **Linguistics:** Social Pragmatics |
| Q-000046 | **Linguistics:** Formal Semantics; Metasemantics<br>**Logic:** Argumentation; General Logic |
| Q-000047 | **Linguistics:** Social Pragmatics<br>**Textuality:** Literary Analysis |
| Q-000048 | **Linguistics:** Linguistic Pragmatics<br>**Textuality:** Philology |
| Q-000049 | **Linguistics:** Discourse Analysis; Formal Semantics; Social Pragmatics |
| Q-000050 | **Linguistics:** Linguistic Pragmatics; Social Pragmatics |
| Q-000051 | **Linguistics:** Linguistic Pragmatics; Social Pragmatics<br>**Logic:** Modal Logic<br>**Textuality:** AI Evaluation |
| Q-000052 | **Linguistics:** Discourse Analysis; Dynamic Semantics; Linguistic Pragmatics |
| Q-000053 | **Linguistics:** Discourse Analysis; Dynamic Semantics; Linguistic Pragmatics; Morphosyntax |
| Q-000054 | **Linguistics:** Lexical Semantics; Linguistic Pragmatics |
| Q-000055 | **Logic:** Conditional Logic; Modal Logic; Probabilistic Logic |
| Q-000056 | **Linguistics:** Discourse Analysis; Dynamic Semantics; Linguistic Pragmatics; Sociolinguistics |
| Q-000057 | **Linguistics:** Formal Semantics; Lexical Semantics; Linguistic Pragmatics |
| Q-000058 | **Linguistics:** Formal Semantics; Lexical Semantics; Linguistic Pragmatics<br>**Textuality:** Literary Analysis |
| Q-000059 | **Linguistics:** Linguistic Pragmatics<br>**Logic:** Conditional Logic; Modal Logic; Probabilistic Logic<br>**Textuality:** Philology |
| Q-000060 | **Linguistics:** Linguistic Pragmatics |
| Q-000061 | **Linguistics:** Dynamic Semantics; Linguistic Pragmatics |
| Q-000062 | **Linguistics:** Linguistic Pragmatics; Social Pragmatics |
| Q-000063 | **Linguistics:** Linguistic Pragmatics<br>**Textuality:** Literary Analysis |
| Q-000064 | **Linguistics:** Discourse Analysis; Lexical Semantics; Linguistic Pragmatics; Morphosyntax<br>**Logic:** Propositional Logic |
| Q-000065 | **Linguistics:** Linguistic Pragmatics; Social Pragmatics; Sociolinguistics |
| Q-000066 | **Linguistics:** Cognitive Pragmatics; Linguistic Pragmatics; Phonology; Social Pragmatics |
| Q-000067 | **Linguistics:** Lexical Semantics; Phonology<br>**Textuality:** Literary Analysis |
| Q-000069 | **Linguistics:** Formal Semantics; Linguistic Pragmatics<br>**Logic:** Epistemology; Modal Logic |
| Q-000070 | **Linguistics:** Discourse Analysis; Social Pragmatics<br>**Textuality:** Philology |
| Q-000071 | **Logic:** Argumentation; Conditional Logic; Propositional Logic |
| Q-000072 | **Linguistics:** Dynamic Semantics; Lexical Semantics<br>**Logic:** Mathematical Intuition |
| Q-000073 | **Linguistics:** Formal Semantics<br>**Logic:** Modal Logic; Propositional Logic |
| Q-000074 | **Linguistics:** Metasemantics<br>**Logic:** Mathematical Intuition |
| Q-000075 | **Textuality:** Literary Analysis; Philology |
| Q-000076 | **Linguistics:** Lexical Semantics<br>**Logic:** Propositional Logic<br>**Textuality:** Literary Analysis |
| Q-000077 | **Linguistics:** Formal Semantics; Lexical Semantics |
| Q-000078 | **Linguistics:** Lexical Semantics; Social Pragmatics<br>**Textuality:** Literary Analysis |
| Q-000079 | **Linguistics:** Social Pragmatics<br>**Textuality:** Literary Analysis |
| Q-000080 | **Linguistics:** Lexical Semantics; Social Pragmatics; Sociolinguistics<br>**Textuality:** Literary Analysis |
| Q-000081 | **Linguistics:** Social Pragmatics; Sociolinguistics<br>**Textuality:** Literary Analysis |
| Q-000082 | **Linguistics:** Formal Semantics; Lexical Semantics<br>**Logic:** Propositional Logic |
| Q-000083 | **Logic:** Epistemic Logic |
| Q-000084 | **Logic:** Argumentation; Conditional Logic; Propositional Logic<br>**Textuality:** AI Evaluation |
| Q-000085 | **Linguistics:** Sociolinguistics<br>**Textuality:** Literary Analysis |
| Q-000087 | **Linguistics:** Dynamic Semantics; Lexical Semantics; Social Pragmatics |
| Q-000088 | **Linguistics:** Discourse Analysis; Lexical Semantics<br>**Logic:** Modal Logic |
| Q-000089 | **Logic:** Argumentation; Modal Logic; Propositional Logic |
| Q-000090 | **Linguistics:** Social Pragmatics<br>**Textuality:** Literary Analysis |
| Q-000091 | **Linguistics:** Morphosyntax<br>**Logic:** Argumentation; Propositional Logic |
| Q-000092 | **Logic:** Argumentation; Modal Logic; Proof Theory |
| Q-000093 | **Logic:** Argumentation; Propositional Logic; Syllogistic Logic |
| Q-000094 | **Linguistics:** Linguistic Pragmatics<br>**Logic:** Argumentation; Propositional Logic |
| Q-000095 | **Logic:** Argumentation; Propositional Logic |
| Q-000096 | **Logic:** Argumentation; Propositional Logic |
| Q-000097 | **Logic:** Argumentation; Propositional Logic |
| Q-000098 | **Linguistics:** Formal Semantics<br>**Logic:** Epistemic Logic |
| Q-000099 | **Linguistics:** Discourse Analysis; Lexical Semantics; Linguistic Pragmatics<br>**Textuality:** Literary Analysis |
| Q-000100 | **Linguistics:** Formal Semantics; Morphosyntax |
| Q-000101 | **Linguistics:** Lexical Semantics; Morphosyntax |
| Q-000102 | **Linguistics:** Formal Semantics; Morphosyntax |
| Q-000103 | **Linguistics:** Discourse Analysis; Lexical Semantics; Linguistic Pragmatics<br>**Textuality:** Literary Analysis |
| Q-000104 | **Linguistics:** Linguistic Pragmatics<br>**Logic:** Epistemic Logic |
| Q-000105 | **Linguistics:** Social Pragmatics; Sociolinguistics<br>**Textuality:** Literary Analysis |
| Q-000106 | **Linguistics:** Formal Semantics; Lexical Semantics |
| Q-000107 | **Linguistics:** Formal Semantics; Lexical Semantics<br>**Logic:** Propositional Logic |
| Q-000108 | **Logic:** Argumentation; Modal Logic; Proof Theory<br>**Textuality:** AI Evaluation |
| Q-000109 | **Linguistics:** Morphosyntax<br>**Logic:** Argumentation; Conditional Logic; Proof Theory |
| Q-000110 | **Linguistics:** Morphosyntax<br>**Logic:** Argumentation; Propositional Logic |
| Q-000111 | **Linguistics:** Morphosyntax<br>**Logic:** Argumentation; Propositional Logic |
| Q-000112 | **Linguistics:** Morphosyntax<br>**Logic:** Argumentation; Propositional Logic<br>**Textuality:** Literary Analysis |
| Q-000113 | **Linguistics:** Morphosyntax<br>**Logic:** Argumentation; Propositional Logic |
| Q-000114 | **Linguistics:** Morphosyntax<br>**Logic:** Argumentation; Propositional Logic |
| Q-000115 | **Logic:** Argumentation; Conditional Logic; Propositional Logic |
| Q-000116 | **Logic:** Argumentation; Conditional Logic; Propositional Logic<br>**Textuality:** AI Evaluation |
| Q-000117 | **Logic:** Argumentation; Propositional Logic |
| Q-000118 | **Linguistics:** Discourse Analysis<br>**Textuality:** Literary Analysis |
| Q-000119 | **Linguistics:** Lexical Semantics; Linguistic Pragmatics<br>**Textuality:** Literary Analysis |
| Q-000120 | **Logic:** Epistemic Logic; Epistemology |
| Q-000121 | **Textuality:** AI Evaluation; Philology |
| Q-000122 | **Textuality:** AI Evaluation |
| Q-000123 | **Textuality:** AI Evaluation; Philology |
| Q-000124 | **Textuality:** AI Evaluation; Philology |
| Q-000127 | **Linguistics:** Linguistic Pragmatics<br>**Textuality:** Literary Analysis; Philology |
| Q-000128 | **Linguistics:** Lexical Semantics; Linguistic Pragmatics<br>**Textuality:** Literary Analysis |
| Q-000129 | **Linguistics:** Formal Semantics<br>**Textuality:** Literary Analysis; Philology |
| Q-000130 | **Linguistics:** Linguistic Pragmatics<br>**Textuality:** Literary Analysis |
| Q-000131 | **Textuality:** Literary Analysis |
| Q-000132 | **Logic:** Argumentation; Propositional Logic |
| Q-000133 | **Logic:** Argumentation; Propositional Logic |
| Q-000134 | **Logic:** Argumentation; Propositional Logic |
| Q-000135 | **Linguistics:** Lexical Semantics<br>**Textuality:** Philology |
| Q-000136 | **Linguistics:** Lexical Semantics<br>**Textuality:** Philology |
| Q-000138 | **Linguistics:** Morphosyntax<br>**Textuality:** Philology |
| Q-000139 | **Linguistics:** Social Pragmatics; Sociolinguistics |
| Q-000140 | **Linguistics:** Discourse Analysis; Linguistic Pragmatics |
| Q-000141 | **Linguistics:** Sociolinguistics |
| Q-000142 | **Linguistics:** Discourse Analysis; Sociolinguistics |
| Q-000143 | **Linguistics:** Discourse Analysis; Sociolinguistics |
| Q-000144 | **Linguistics:** Discourse Analysis; Social Pragmatics; Sociolinguistics |
| Q-000145 | **Linguistics:** Social Pragmatics; Sociolinguistics |
| Q-000146 | **Linguistics:** Social Pragmatics; Sociolinguistics |
| Q-000147 | **Linguistics:** Social Pragmatics; Sociolinguistics |
| Q-000148 | **Linguistics:** Social Pragmatics; Sociolinguistics |
| Q-000149 | **Linguistics:** Social Pragmatics; Sociolinguistics |
| Q-000150 | **Linguistics:** Dynamic Semantics; Linguistic Pragmatics; Social Pragmatics |
| Q-000151 | **Linguistics:** Discourse Analysis; Social Pragmatics; Sociolinguistics |
| Q-000152 | **Linguistics:** Discourse Analysis; Linguistic Pragmatics; Social Pragmatics |
| Q-000153 | **Linguistics:** Discourse Analysis; Social Pragmatics; Sociolinguistics |
| Q-000154 | **Linguistics:** Social Pragmatics; Sociolinguistics<br>**Textuality:** Literary Analysis |
| Q-000155 | **Linguistics:** Morphosyntax; Sociolinguistics |
| Q-000156 | **Linguistics:** Morphosyntax; Sociolinguistics |
| Q-000157 | **Linguistics:** Morphosyntax; Social Pragmatics; Sociolinguistics |
| Q-000158 | **Linguistics:** Sociolinguistics |
| Q-000159 | **Linguistics:** Sociolinguistics |
| Q-000160 | **Linguistics:** Social Pragmatics; Sociolinguistics |
| Q-000161 | **Linguistics:** Discourse Analysis; Social Pragmatics |
| Q-000162 | **Linguistics:** Discourse Analysis; Linguistic Pragmatics |
| Q-000163 | **Linguistics:** Discourse Analysis; Social Pragmatics |
| Q-000164 | **Linguistics:** Lexical Semantics; Social Pragmatics<br>**Textuality:** Literary Analysis |
| Q-000165 | **Linguistics:** Discourse Analysis; Linguistic Pragmatics |
| Q-000166 | **Linguistics:** Lexical Semantics; Linguistic Pragmatics |
| Q-000167 | **Textuality:** AI Evaluation |
| Q-000168 | **Textuality:** AI Evaluation; Philology |
| Q-000169 | **Logic:** Epistemic Logic |
| Q-000170 | **Logic:** Argumentation; Conditional Logic; Propositional Logic |
| Q-000171 | **Linguistics:** Formal Semantics<br>**Logic:** Epistemology |
| Q-000172 | **Logic:** Modal Logic |
| Q-000173 | **Logic:** Modal Logic |
| Q-000174 | **Linguistics:** Formal Semantics<br>**Logic:** Argumentation; Epistemic Logic; Propositional Logic |
| Q-000175 | **Logic:** Argumentation; Modal Logic<br>**Textuality:** AI Evaluation |
| Q-000176 | **Logic:** Modal Logic; Propositional Logic |
| Q-000177 | **Logic:** Modal Logic |
| Q-000178 | **Logic:** Modal Logic; Propositional Logic |
| Q-000179 | **Linguistics:** Formal Semantics<br>**Logic:** Argumentation; Epistemic Logic; Propositional Logic |
| Q-000180 | **Logic:** Syllogistic Logic |
| Q-000181 | **Logic:** Syllogistic Logic |
| Q-000182 | **Linguistics:** Formal Semantics<br>**Logic:** Predicate Logic; Propositional Logic |
| Q-000183 | **Linguistics:** Formal Semantics<br>**Logic:** Argumentation; Propositional Logic |
| Q-000184 | **Linguistics:** Formal Semantics; Morphosyntax<br>**Logic:** Argumentation |
| Q-000185 | **Logic:** Modal Logic |
| Q-000186 | **Linguistics:** Morphosyntax<br>**Logic:** Modal Logic |
| Q-000187 | **Linguistics:** Formal Semantics<br>**Logic:** Argumentation; Predicate Logic |
| Q-000188 | **Linguistics:** Formal Semantics<br>**Logic:** Predicate Logic |
| Q-000189 | **Linguistics:** Formal Semantics<br>**Logic:** Argumentation; Propositional Logic<br>**Textuality:** AI Evaluation |
| Q-000190 | **Linguistics:** Formal Semantics<br>**Logic:** Argumentation |
| Q-000191 | **Linguistics:** Formal Semantics<br>**Logic:** Propositional Logic |
| Q-000192 | **Linguistics:** Formal Semantics<br>**Logic:** Argumentation; Propositional Logic |
| Q-000193 | **Linguistics:** Formal Semantics<br>**Logic:** Argumentation; Predicate Logic |
| Q-000194 | **Linguistics:** Cognitive Pragmatics; Formal Semantics; Linguistic Pragmatics |
| Q-000195 | **Linguistics:** Formal Semantics; Lexical Semantics<br>**Textuality:** AI Evaluation |
| Q-000196 | **Linguistics:** Formal Semantics; Lexical Semantics; Linguistic Pragmatics<br>**Textuality:** AI Evaluation |
| Q-000197 | **Linguistics:** Formal Semantics; Lexical Semantics; Linguistic Pragmatics<br>**Textuality:** AI Evaluation |
| Q-000198 | **Logic:** Syllogistic Logic<br>**Textuality:** AI Evaluation |
| Q-000199 | **Logic:** Syllogistic Logic<br>**Textuality:** AI Evaluation |
| Q-000200 | **Logic:** Syllogistic Logic<br>**Textuality:** AI Evaluation |
| Q-000201 | **Logic:** Syllogistic Logic<br>**Textuality:** AI Evaluation |
| Q-000202 | **Linguistics:** Metasemantics<br>**Logic:** Epistemology |
| Q-000203 | **Linguistics:** Metasemantics<br>**Logic:** Epistemology |
| Q-000204 | **Linguistics:** Metasemantics<br>**Logic:** Epistemology |
| Q-000205 | **Linguistics:** Morphosyntax; Social Pragmatics<br>**Logic:** Epistemology |
| Q-000206 | **Linguistics:** Social Pragmatics<br>**Textuality:** Literary Analysis |
| Q-000207 | **Linguistics:** Formal Semantics; Lexical Semantics; Morphosyntax |
| Q-000208 | **Linguistics:** Morphosyntax<br>**Textuality:** Philology |
| Q-000209 | **Linguistics:** General Linguistics; Lexical Semantics; Morphosyntax |
| Q-000210 | **Linguistics:** Formal Semantics; General Linguistics; Morphosyntax |
| Q-000211 | **Linguistics:** General Linguistics; Morphosyntax |
| Q-000212 | **Linguistics:** General Linguistics; Morphosyntax |
| Q-000213 | **Linguistics:** General Linguistics; Morphosyntax |
| Q-000214 | **Linguistics:** Discourse Analysis |
| Q-000215 | **Linguistics:** Linguistic Pragmatics; Social Pragmatics<br>**Textuality:** Literary Analysis |
| Q-000216 | **Linguistics:** Sociolinguistics<br>**Textuality:** Philology |
| Q-000218 | **Linguistics:** Lexical Semantics<br>**Textuality:** Philology |
| Q-000219 | **Linguistics:** Linguistic Pragmatics<br>**Textuality:** AI Evaluation |
| Q-000220 | **Linguistics:** Formal Semantics; Lexical Semantics; Linguistic Pragmatics<br>**Textuality:** AI Evaluation |
| Q-000221 | **Linguistics:** Formal Semantics; Linguistic Pragmatics |
| Q-000222 | **Linguistics:** Discourse Analysis; Formal Semantics; Linguistic Pragmatics |
| Q-000223 | **Linguistics:** Social Pragmatics; Sociolinguistics |
| Q-000225 | **Linguistics:** Morphosyntax<br>**Textuality:** Philology |
| Q-000226 | **Linguistics:** Formal Semantics; Morphosyntax |
| Q-000227 | **Linguistics:** Formal Semantics; Linguistic Pragmatics |
| Q-000228 | **Linguistics:** General Linguistics; Phonology<br>**Textuality:** Philology |
| Q-000229 | **Linguistics:** Phonology<br>**Textuality:** Philology |
| Q-000230 | **Linguistics:** Phonology<br>**Textuality:** Philology |
| Q-000231 | **Linguistics:** Phonology<br>**Textuality:** Philology |
| Q-000232 | **Linguistics:** Morphosyntax; Phonology<br>**Textuality:** Philology |
| Q-000233 | **Linguistics:** Phonology<br>**Textuality:** Philology |
| Q-000234 | **Linguistics:** Phonology<br>**Textuality:** Philology |
| Q-000235 | **Linguistics:** Phonology<br>**Textuality:** Philology |
| Q-000236 | **Linguistics:** Phonology<br>**Textuality:** Philology |
| Q-000237 | **Linguistics:** Phonology<br>**Textuality:** Philology |
| Q-000238 | **Linguistics:** Phonology<br>**Textuality:** Philology |
| Q-000239 | **Linguistics:** Phonology<br>**Textuality:** Philology |
| Q-000240 | **Linguistics:** Phonology<br>**Textuality:** Philology |
| Q-000241 | **Linguistics:** Phonology<br>**Textuality:** Philology |
| Q-000242 | **Linguistics:** Phonology<br>**Textuality:** Philology |
| Q-000243 | **Linguistics:** Phonology<br>**Textuality:** Philology |
| Q-000244 | **Linguistics:** Phonology<br>**Textuality:** Philology |
| Q-000245 | **Linguistics:** Phonology<br>**Textuality:** Philology |
| Q-000246 | **Linguistics:** Phonology<br>**Textuality:** Philology |
| Q-000247 | **Linguistics:** Phonology<br>**Textuality:** Philology |
| Q-000248 | **Linguistics:** Morphosyntax<br>**Textuality:** Philology |
| Q-000249 | **Linguistics:** Morphosyntax<br>**Textuality:** Philology |
| Q-000250 | **Linguistics:** Morphosyntax; Phonology<br>**Textuality:** Philology |
| Q-000251 | **Linguistics:** Morphosyntax<br>**Textuality:** Literary Analysis; Philology |
| Q-000252 | **Linguistics:** Morphosyntax<br>**Textuality:** Literary Analysis; Philology |
| Q-000253 | **Linguistics:** Morphosyntax<br>**Textuality:** Philology |
| Q-000255 | **Linguistics:** Phonology<br>**Textuality:** Philology |
| Q-000256 | **Linguistics:** Phonology<br>**Textuality:** Philology |